\documentclass[]{phai}
\usepackage{amsmath,amsfonts,bm}

\def\eqref#1{equation~\ref{#1}}
\def\1{\bm{1}}

\DeclareMathAlphabet{\mathsfit}{\encodingdefault}{\sfdefault}{m}{sl}
\SetMathAlphabet{\mathsfit}{bold}{\encodingdefault}{\sfdefault}{bx}{n}

\usepackage{url}
\usepackage{tabularx,array,colortbl}
\usepackage{makecell}
\usepackage{amssymb}
\usepackage{bookmark}
\usepackage{needspace}

\hypersetup{
  pdftitle={TouchScale: 500 Hours of Human Vision and Touch for Visual-Tactile Learning},
  pdfauthor={Dayou Li; Hao Wang; Qianqian Yang; Zihao Zhu; Haoquan Fang; Ziyao Zeng; Yan Han; Zihan Wang; Yan Wang; Baoru Huang; Dilin Wang; Kenji Shimada; Yiyue Luo; Manling Li; Teresa Ly; Mustafa Mukadam; Rakesh Ranjan; Ruohan Zhang; Qi He; Changliu Liu; Xu Chen; Marco Pavone; Bangya Liu; Jiachen Li; Masayoshi Tomizuka; Zhiwen Fan}
}

\title{TouchScale: 500 Hours of Human Vision and Touch for Visual-Tactile Learning}

\author[1,*]{Dayou~Li}
\author[2,*]{Hao~Wang}
\author[3,*]{Qianqian~Yang}
\author[1,*]{Zihao~Zhu}
\author[4]{Haoquan~Fang}
\author[5]{Ziyao~Zeng}
\author[6]{Yan~Han}
\author[7]{Zihan~Wang}
\author[8]{Yan~Wang}
\author[9]{Baoru~Huang}
\author[10]{Dilin~Wang}
\author[3]{Kenji~Shimada}
\author[11]{Yiyue~Luo}
\author[12]{Manling~Li}
\author[13]{Teresa~Lv}
\author[11]{Mustafa~Mukadam}
\author[10]{Rakesh~Ranjan}
\author[4]{Ruohan~Zhang}
\author[6]{Qi~He}
\author[3]{Changliu~Liu}
\author[11]{Xu~Chen}
\author[4,8]{Marco~Pavone}
\author[7]{Bangya~Liu}
\author[14]{Jiachen~Li}
\author[15]{Masayoshi~Tomizuka}
\author[1,\dagger]{Zhiwen~Fan}

\affiliation[1]{Texas A\&M University}
\affiliation[2]{Google DeepMind}
\affiliation[3]{CMU}
\affiliation[4]{Stanford University}
\affiliation[5]{Yale University}
\affiliation[6]{Microsoft}
\affiliation[7]{Overfit Lab}
\affiliation[8]{NVIDIA}
\affiliation[9]{University of Liverpool}
\affiliation[10]{Meta}
\affiliation[11]{University of Washington}
\affiliation[12]{Northwestern University}
\affiliation[13]{Sony}
\affiliation[14]{Georgia Tech}
\affiliation[15]{UC Berkeley}

\contribution[*]{Equal contribution}
\contribution[\dagger]{Corresponding author}

\abstract{
Large-scale egocentric human interaction data is becoming an important source of physical supervision for embodied learning, yet video alone leaves the contact and pressure that characterize physical interaction unrecorded. Recent visual--tactile datasets provide this missing supervision, but their synchronized tactile data remain far smaller in volume than human video. Moreover, the largest resources often merge recordings from different sensors or annotation procedures, which makes the effect of data scale difficult to isolate. In short, datasets that capture hundreds of hours of human vision and touch through one consistent sensing pipeline remain scarce. We therefore introduce TouchScale, a 500-hour dataset of contact-rich human interaction recorded with a single unified wearable setup. Its approximately 2K predefined task descriptions span everyday activities and structured manipulation, and each recording temporally aligns egocentric RGB-D video with wrist RGB video and dense full-hand bimanual tactile measurements. With TouchScale, we ask what scaling human visual--tactile data unlocks for perception and robot learning. Compared with prior tactile data, training on the full TouchScale raises zero-shot contact IoU on data from an unseen tactile sensor from 0.134 to 0.383. Pretraining a visual encoder on TouchScale also yields the highest action recognition accuracy on three benchmarks among the compared visual--tactile datasets. Used for visual--tactile mid-training of a robot policy, TouchScale improves the average real-world success rate across four contact-rich manipulation tasks from 22.5\% to 57.5\%. With the sensor and collection protocol held fixed, both zero-shot tactile prediction and robot success show an overall upward trend as more TouchScale data is used. These results suggest that human visual--tactile data collected at scale with consistent sensing benefits both perception and robot manipulation. We will publicly release TouchScale, including all synchronized visual–tactile recordings and reconstructed object models, to support future research on scalable visual–tactile learning.
}
\date{September 27, 2026}
\metadata[Keywords]{Visual-Tactile Learning, Egocentric Dataset, Robotic Manipulation}
\metadata[Project Website]{\url{https://touch-scale.github.io/}}
\correspondence{Zhiwen Fan: \email{zhiwenfan@tamu.edu}}

\begin{document}
\maketitle
\begin{figure*}[!htbp]
\centering
\includegraphics[width=\textwidth,keepaspectratio,clip]{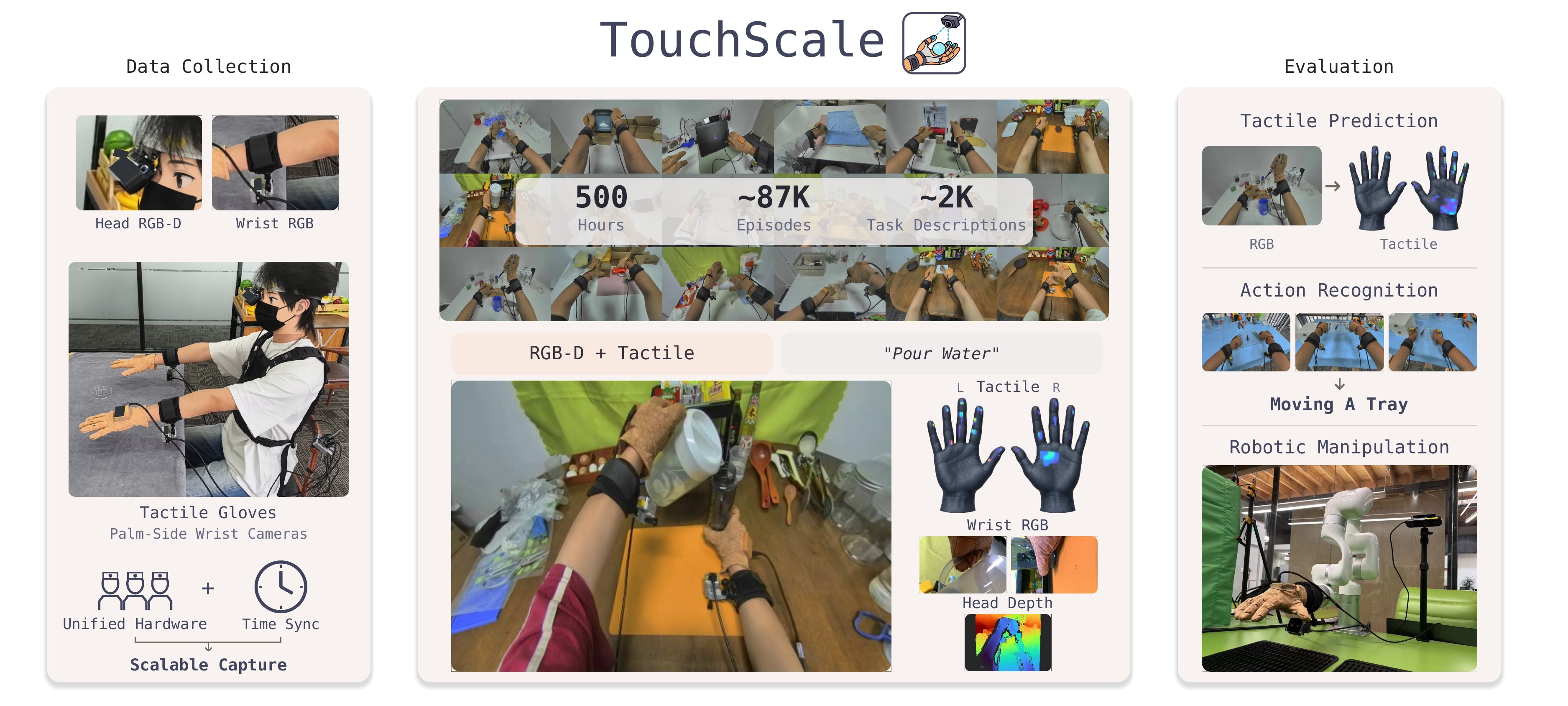}
\caption{\textbf{Overview.}
TouchScale is a large-scale human visual--tactile dataset collected with a unified wearable system.
\textbf{Capture:} Human interactions are recorded with synchronized head RGB-D, wrist RGB from both hands, and bimanual tactile measurements.
\textbf{Dataset:} TouchScale contains 500 hours of data, approximately 87K episodes, and approximately 2K task descriptions spanning diverse everyday interactions.
\textbf{Evaluation:} We evaluate it on tactile prediction, action recognition, and robotic manipulation.}
\label{fig:teaser}
\end{figure*}

\section{Introduction}
When a person lifts an egg or wrings a wet cloth, the outcome depends on where the fingers press and how hard, yet a camera watching the hand records only its motion. Similarly, egocentric video datasets now span hundreds to thousands of hours of human activity, from everyday tasks to fine-grained hand-object manipulation \citep{grauman2022ego4d,grauman2024egoexo4d,hoque2026egodex}, and have begun to support robot learning \citep{kareer2025egomimic,zheng2026egoscale}, yet they capture how hands move while leaving contact and pressure unrecorded. The missing signal is increasingly important for robotics, as tactile sensing has become a common input for contact-rich manipulation \citep{yin2023touchdexterity,yuan2024robotsynesthesia,huang2025vitac,yu2025mimictouch,liu2025vtdexmanip,niu2026trex}. Existing egocentric visual--tactile datasets provide this signal, but most contain fewer than 30 hours of recorded interaction \citep{song2025opentouch,dessalene2026feel,zhou2026touchanything}. Thus the effect of data scale is difficult to study within any single one, and pooling several datasets introduces a second difficulty, as their differences in sensing hardware and collection protocol confound comparisons across them. Although recent studies using egocentric human data have reported favorable scaling trends for downstream robot learning \citep{kareer2025egomimic,zhu2026emma,punamiya2026egoverse,zheng2026egoscale}, it remains unclear whether these benefits extend to visual--tactile learning and robotic manipulation when tactile data are collected at the scale of hundreds of hours under a unified sensing pipeline.

To address this gap, we introduce \textbf{TouchScale}, a 500-hour visual--tactile dataset that pairs diverse human interaction with consistent multimodal sensing from a unified wearable capture pipeline. Each recording synchronizes egocentric RGB-D video with wrist RGB videos and dense full-hand tactile measurements from both hands, and the recordings span everyday activities and structured manipulation tasks with diverse hand motions and interaction patterns. Using the same wearable setup, we record interactions under approximately 2K task descriptions with more than 1.5K objects, spanning 9 high-level collection settings that contain over 800 distinct scene configurations. All recordings share the same generation of sensing hardware and synchronization pipeline, so visual and tactile measurements remain consistent throughout the dataset. The wearable setup can also be deployed repeatedly across participants and environments without requiring manual action annotation, which offers a practical collection paradigm to further scale human visual--tactile data.

With TouchScale, we study four questions about large-scale human visual--tactile data. First, we ask whether training on TouchScale enables zero-shot vision-to-touch prediction on a tactile sensor unseen during training. With the same prediction architecture, the model trained on a size-matched 16-hour subset of TouchScale reaches a contact IoU (cIoU) of 0.181 on EgoTactile, compared with 0.134 for the one trained on the 16-hour EgoTouch training split \citep{zhou2026touchanything}. Second, we ask whether tactile supervision on TouchScale learns visual representations that transfer beyond tactile prediction, and pretraining on TouchScale yields higher action recognition accuracy than pretraining on the compared visual--tactile datasets across three benchmarks under both linear probing and end-to-end fine-tuning. Third, we ask whether human visual--tactile data improves robot policies. Prior work has used tactile observations directly in robot policies or combined human tactile data with action supervision for transfer to robots \citep{liu2025vtdexmanip,niu2026trex,zhang2026human}, whereas we use TouchScale only for visual--tactile mid-training of $\mathcal{N}_0$-VTLA \citep{n0vtla2026} without retargeting human hand motion to the robot. This mid-training raises the average real-world success rate on four contact-rich manipulation tasks from 22.5\% to 57.5\%. Finally, we ask whether these gains grow with data scale when the sensor and collection protocol stay fixed. Increasing the training data from one tenth of TouchScale to the full dataset raises zero-shot cIoU on EgoTactile from 0.311 to 0.383, and growing the mid-training data from one fifth to the full dataset lifts robot success from 30.0\% to 57.5\%. Together, these results indicate that large-scale human visual--tactile data benefits embodied learning from perception to robot control, with gains that tend to increase over the range of scales we test.

\Needspace{6\baselineskip}
In summary, our main contributions are:
\begin{itemize}
    \item We introduce TouchScale, a 500-hour human visual--tactile dataset spanning diverse physical interactions, together with a scalable wearable capture paradigm that combines egocentric RGB-D and bimanual wrist RGB cameras with dense full-hand tactile gloves, and we will make the dataset publicly available.
    
    \item We evaluate TouchScale from perception to robot control. Models trained on TouchScale transfer zero-shot to an unseen tactile sensor, and tactile-supervised pretraining on TouchScale yields visual representations that transfer to action recognition. Used as mid-training supervision without human-to-robot action alignment, TouchScale raises the average real-world success rate across four contact-rich manipulation tasks from 22.5\% to 57.5\%.
    
    \item We study data scaling with the sensor and collection protocol held fixed, varying only the amount of training data. Both zero-shot tactile prediction and real-world manipulation success show an overall upward trend as TouchScale data increases, suggesting that the scaling benefits reported for egocentric video also hold for visual--tactile learning.
\end{itemize}

\section{Related Work}

\begin{table*}[!tbp]
    \centering
    \footnotesize
    \setlength{\tabcolsep}{2.5pt}
    \renewcommand{\arraystretch}{1.13}
    \resizebox{\linewidth}{!}{%
    \begin{tabular}{l*{10}{c}}
    \toprule
    Dataset & Hours & Samples & Vision & Views & Hands & Sensor & Components & Taxels/hand & Coverage & Objects \\
    \midrule

    \shortstack[l]{STAG\\\cite{sundaram2019stag}}
    & NR & 135K frames & -- & -- & Single & Glove & Normal & 548 & Hand & 26 \\

    \shortstack[l]{ActionSense\\\cite{delpreto2022actionsense}}
    & 12.97 & NR & RGB-D & H+6Exo & Both & Glove & Normal & 682 & \shortstack{Hand excl.\\tips} & 21 \\

    \shortstack[l]{PressureVisionDB\\\cite{grady2022pressurevision}}
    & 16 & $\sim$3M frames & RGB & 4Exo & Single & Pad & Normal & -- & Surface & -- \\

    \shortstack[l]{ContactLabelDB\\\cite{grady2024pressurevisionpp}}
    & NR & \shortstack{500K pressure\\frames} & RGB & Exo & Single & Pad & Normal & -- & Tips & -- \\

    \shortstack[l]{EgoPressure\\\cite{zhao2025egopressure}}
    & 5 & 4.3M frames & RGB-D & H+7Exo & Single & Pad & Normal & -- & Surface & -- \\

    \shortstack[l]{OpenTouch\\\cite{song2025opentouch}}
    & 5.1 & 2.9K clips & RGB & H & Single & Glove & Normal & 169 & Hand & $\sim$800 types \\

    \shortstack[l]{FEEL\\\cite{dessalene2026feel}}
    & $\sim$27 & 3M frames & RGB & H & Both & Glove & Normal & 6 & Tips+palm & NR \\

    \shortstack[l]{EgoTouch\\\cite{zhou2026touchanything}}
    & 20 & 2.1M frames & RGB & H+2W & Both & Glove & Normal & 16$\times$16 & Palm & 1000+ \\

    \shortstack[l]{EgoTactile\\\cite{zeng2026egotactile}}
    & 5.82 & 768 clips & RGB & H/neck & Single & Glove & Normal & 162 & Hand & 63 \\

    \shortstack[l]{DeskTask-Tac\\\cite{zhang2026human}}
    & 37.2 & $\sim$4M frames & RGB & H+2Exo & Both & Glove & NR & NR & Hand & NR \\

    % HumanTouch~\cite{humantouch2026}
    % & $\sim$100 & 13,469 episodes & RGB & H+2W & Both & Glove & Normal & 360 & Hand & NR \\

    \rowcolor[gray]{0.935}
    \textbf{TouchScale}
    & \textbf{500} & \textbf{$\sim$87K episodes} & \textbf{RGB-D} & \textbf{H+2W} & \textbf{Both} & \textbf{Glove} & \textbf{Normal} & \textbf{880} & \textbf{Hand} & \textbf{1.5k+} \\

    \bottomrule
    \end{tabular}%
    }
\caption{\textbf{Comparison of human tactile and visual--tactile datasets.}
H, W, and Exo denote head, wrist, and external views, respectively; NR denotes not reported and -- denotes not applicable. We report the pressure-labeled subset of ContactLabelDB and the directly recorded human subset of H-Tac as DeskTask-Tac. Among these datasets, TouchScale offers the most recorded hours and the densest glove-based tactile sensing.}
\label{tab:visual_tactile_datasets}
\end{table*}

\subsection{Egocentric Visual--Tactile Datasets}

Large-scale egocentric human interaction data is becoming an important source of supervision for embodied learning. Existing datasets cover daily activities and manipulation~\citep{damen2022epickitchens100,grauman2022ego4d,grauman2024egoexo4d,hoque2026egodex,liu2026egotl,punamiya2026egoverse,robotics2025world,zhang2026egotac}, with recent collections exceeding 20,000 hours~\citep{zheng2026egoscale}, but generally lack synchronized tactile measurements. EgoPressure, EgoTactile, and EgoTouch study pressure prediction from egocentric or multi-view video, covering hand--surface contact, object grasping, and bimanual manipulation~\citep{zhao2025egopressure,zeng2026egotactile,zhou2026touchanything}, while TouchSight introduces TwinTouch-20H by pairing glove recordings with generated bare-hand videos that share the measured tactile labels~\citep{zhou2026touchsight}. OpenTouch and HT-Bench evaluate tactile representations and cross-modal retrieval, with HT-Bench also studying masked tactile modeling and vision-to-tactile synthesis~\citep{song2025opentouch,huang2026htbench}, while FEEL studies force supervision for physical interaction understanding and visual representation learning~\citep{dessalene2026feel}. DeskTask-Tac contains 37.2 h of human visual--tactile recordings~\citep{zhang2026human}. Table~\ref{tab:visual_tactile_datasets} compares dataset scale and sensing configurations. TouchScale contains 500 hours collected through a consistent sensing pipeline, with egocentric RGB-D, wrist RGB, and bimanual tactile measurements. We study how data scale affects cross-dataset tactile prediction, transferable visual representations, and robot learning.

\subsection{Human Visual--Tactile Data for Robot Learning}

Human video has been used for visual pretraining and human--robot policy learning~\citep{nair2023r3m,ma2023vip,kareer2025egomimic,qiu2025humanoid,yang2025egovla}. Tactile recordings provide direct measurements of contact and pressure, but differences in sensor layout and hand geometry complicate transfer from human hands to robots. Touch in the Wild uses human demonstrations collected with a portable tactile gripper to pretrain visual--tactile representations for downstream manipulation~\citep{zhu2025touchwild}. TactAlign and TTP study human--robot transfer through shared tactile representations: TactAlign aligns observations across human and robot sensors, while TTP pretrains on mixed human and robot interaction data using future tactile prediction~\citep{wi2026tactalign,zhang2026human}. Tactile supervision is also used during policy mid-training. T-Rex introduces tactile-rich robot mid-training after large-scale human pretraining without tactile measurements~\citep{niu2026trex}, whereas $N_0$-VTLA trains a future-tactile predictor on robot data before aligning its latent tokens with an action expert~\citep{n0vtla2026}. We use TouchScale for vision-to-tactile mid-training on human recordings without human action labels. The human-data stage does not require retargeting hand motion or matching human and robot action spaces. We then evaluate whether this supervision improves real-robot task success after robot-specific training.
\section{TouchScale Dataset}
\label{sec:dataset}

\begin{figure*}[t]
    \centering
    \includegraphics[width=\textwidth,keepaspectratio,clip]{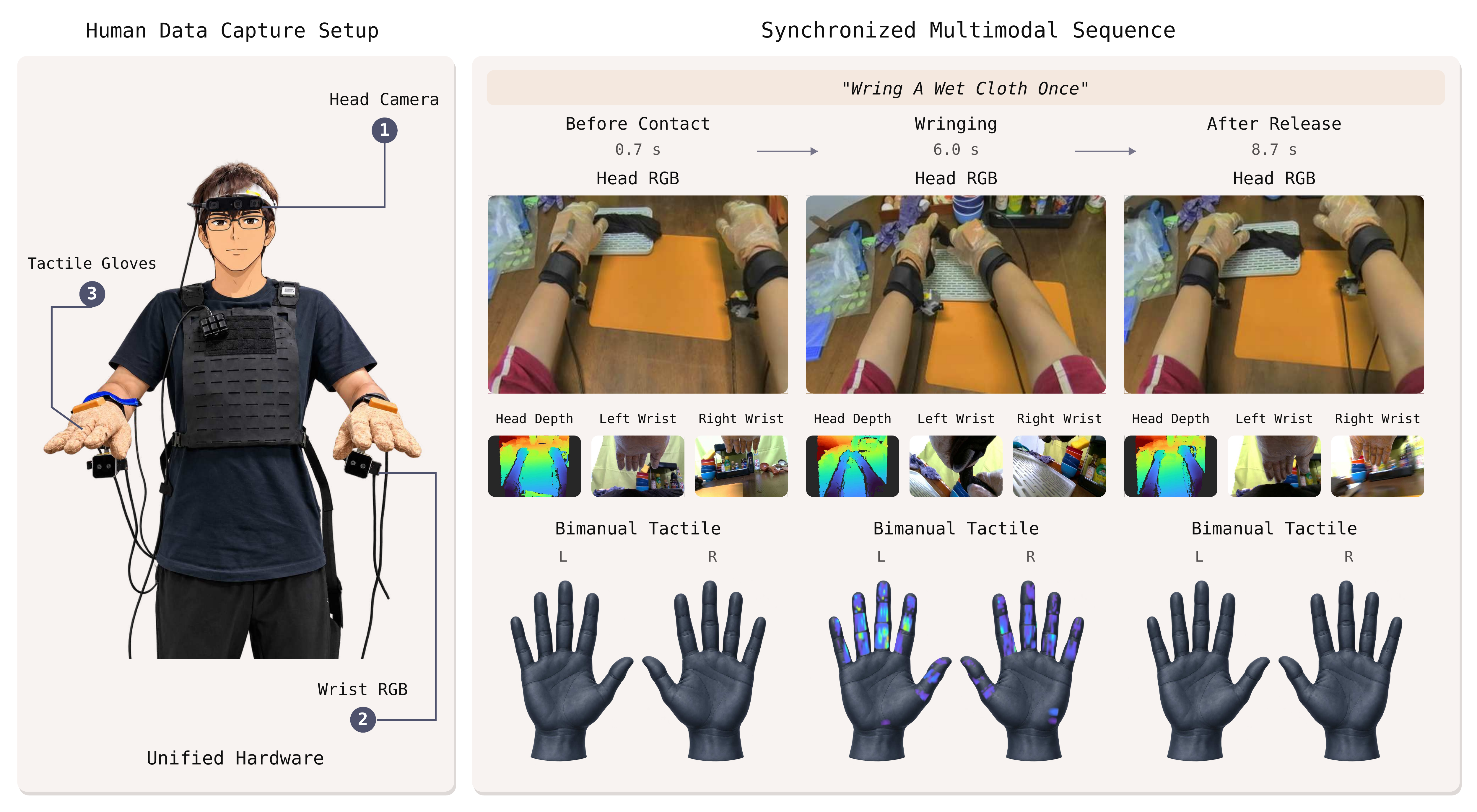}
    \caption{
    \textbf{TouchScale data collection.}
    Left: the unified wearable setup used throughout the collection, consisting of a head-mounted RGB-D camera, two wrist-mounted RGB cameras, and bimanual tactile gloves.
    Right: time-aligned observations from an example interaction, including head RGB-D, bimanual wrist RGB, and bimanual tactile measurements at different stages of the task.
    }
    \label{fig:data_collection}
\end{figure*}

TouchScale is a 500-hour human visual--tactile dataset collected with a unified wearable sensing setup. Each recording contains egocentric RGB-D video, bimanual wrist RGB video, and bimanual tactile measurements. Fig.~\ref{fig:data_collection} presents the capture setup and multimodal observations, and Fig.~\ref{fig:data_diversity} summarizes the task, scene, and interaction coverage of the dataset.

\subsection{Data Collection Setup}
\label{sec:collection_setup}

TouchScale uses a unified wearable setup for human visual--tactile data collection, as illustrated in Fig.~\ref{fig:data_collection}. The setup combines an egocentric RGB-D camera, two wrist-mounted RGB cameras, and bimanual tactile gloves.

\paragraph{Egocentric and wrist cameras.}
As shown in Fig.~\ref{fig:data_collection}, we use an Orbbec Gemini 345Lg stereo RGB-D camera as the egocentric sensor, recording RGB at $1280\times720$ and depth at $640\times480$, both at 30\,Hz. The depth stream provides metric 3D scene geometry from the egocentric view throughout the interaction. Two Intel RealSense D405 cameras are mounted on the wrists and record RGB at $1280\times720$ and 30\,Hz. The egocentric camera captures the overall interaction, while the wrist cameras provide close-up views of hand--object interaction and complement the egocentric view when contact is partially occluded.

\paragraph{Tactile gloves.}
We use flexible tactile sensing gloves to record bimanual tactile signals during interaction, as shown in Fig.~\ref{fig:data_collection}. Each glove contains 880 sensing taxels distributed across the five fingers and palm, with a spatial resolution below 2\,mm. The dense sensing layout provides hand-wide coverage of contact during manipulation. The gloves record normal tactile pressure across the hand. Additional hardware specifications and the sensor layout are provided in Appendix~\ref{app:tactile_glove}.

\subsection{Data Collection and Task Design}
\label{sec:collection_protocol}

\begin{figure*}[t]
    \centering
    \includegraphics[width=\textwidth,keepaspectratio,clip]{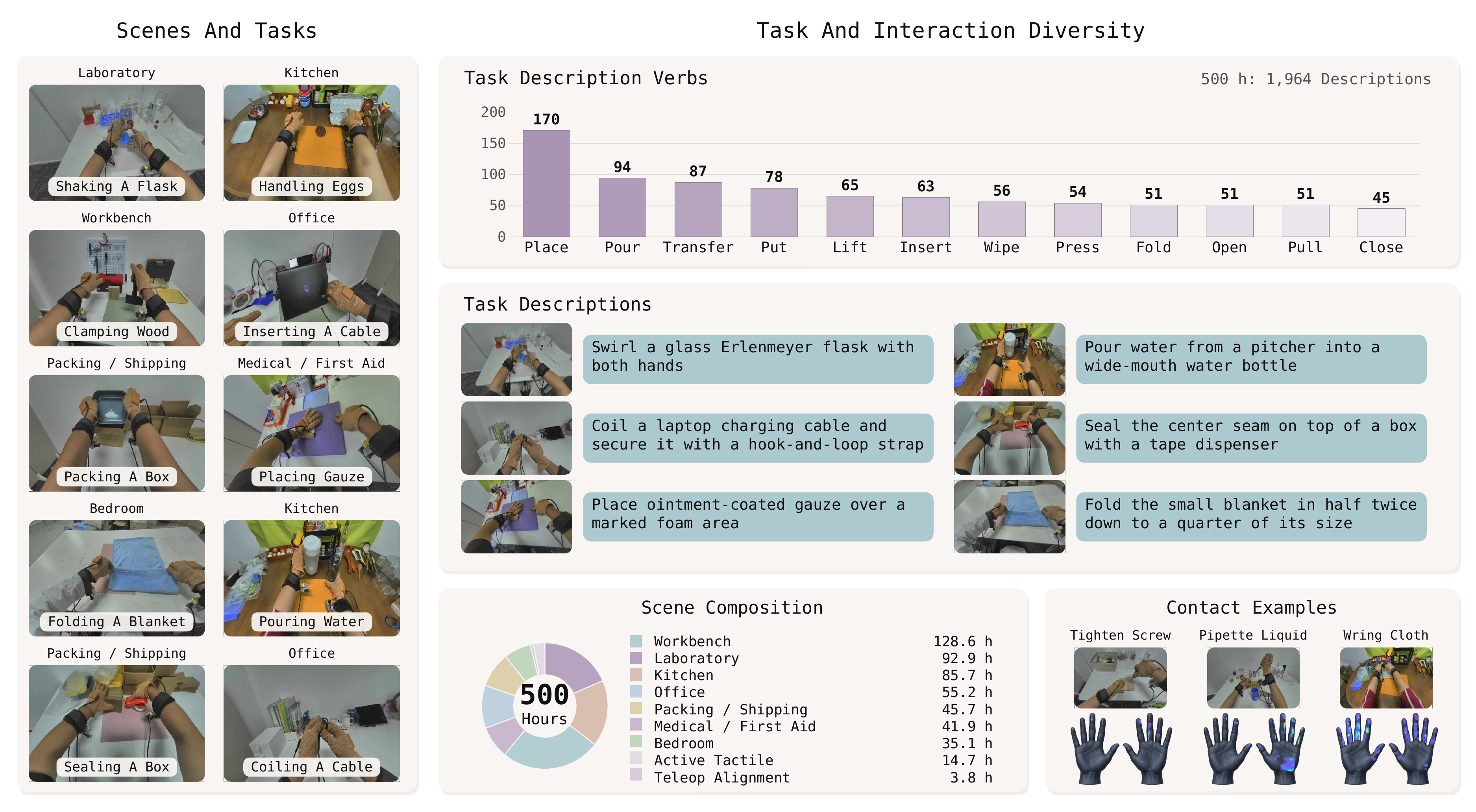}
    \caption{\textbf{Task and interaction diversity in TouchScale.}
Left: representative tasks across different collection settings.
Right: task-description verb frequencies, example task instructions, the collection-setting composition of the 500-hour dataset, and representative tactile contact patterns.}
    \label{fig:data_diversity}
\end{figure*}

\paragraph{Diverse everyday activities.}
TouchScale contains approximately 2K task descriptions across 9 high-level collection settings and 800+ distinct scene configurations, including laboratory, kitchen, workbench, office, bedroom, medical, and packing environments. Laboratory, kitchen, and workbench activities account for the largest portions of the collection. The tasks cover transport and placement, pouring and transfer, insertion and alignment, tool use, opening and closing, folding, wiping, pressing, and other contact-rich interactions, including both single-hand and bimanual manipulation. As shown in Fig.~\ref{fig:data_diversity}, \emph{place} is the most frequent task verb, followed by \emph{pour}, \emph{transfer}, and \emph{put}, with many other interaction types represented throughout the dataset. Each task is written as a natural-language instruction with an explicit action, the manipulated object or objects, and the intended interaction or target state. When needed, the instruction further specifies the tool, target location, execution order, hand use, or completion condition. Representative examples in Fig.~\ref{fig:data_diversity} include ``Swirl a glass Erlenmeyer flask with both hands,'' ``Pour water from a pitcher into a wide-mouth water bottle,'' ``Coil a laptop charging cable and secure it with a hook-and-loop strap,'' and ``Seal the center seam on top of a box with a tape dispenser.''

\paragraph{Diverse objects, poses, and behaviors.}
The 500-hour collection includes 1.5k+ objects spanning different geometries, materials, and physical properties. For repeated recordings of a task, we vary object instances and initial configurations, including object position, orientation, relative arrangement, and scene layout. Approximately 20 participants contribute to the collection, producing differences in grasp choice, movement timing, hand coordination, contact sequence, and motion trajectory even under the same task instruction. These variations lead to different hand--object poses and tactile contact patterns across repeated executions. Fig.~\ref{fig:data_diversity} shows distinct contact distributions for tightening a screw, pipetting liquid, and wringing a cloth, while Fig.~\ref{fig:data_collection} shows how contact evolves as a wet cloth is grasped, wrung, and released. We additionally provide reconstructed 3D objects from object-centric RGB captures, and the procedure is described in Appendix~\ref{app:object_reconstruction}.

\begin{figure}[t]
    \centering
    \includegraphics[width=\linewidth]{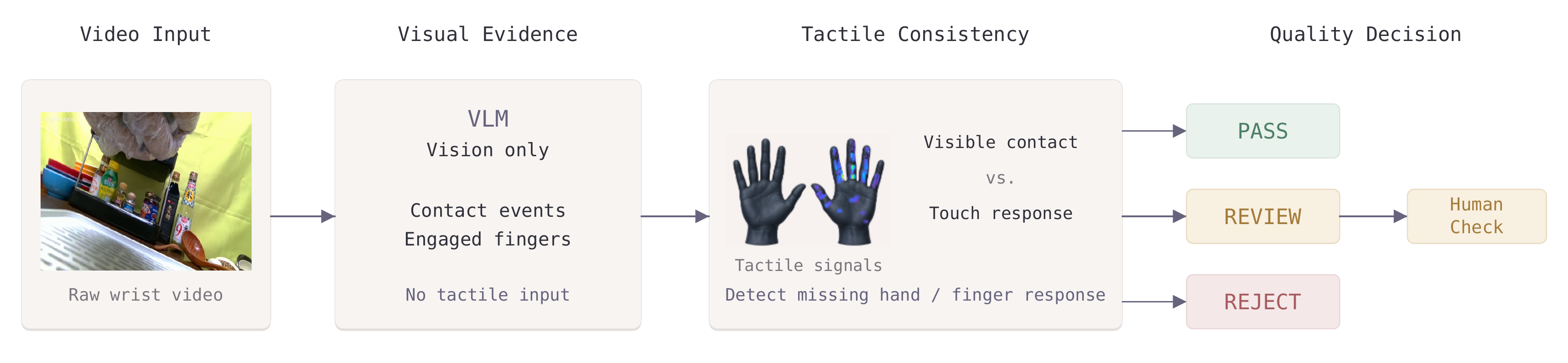}
    \caption{\textbf{Postprocessing for data quality.} Contact events and engaged fingers are first identified from wrist video without tactile input, then compared with tactile responses to detect missing signals. The resulting consistency check assigns PASS, REVIEW, or REJECT, with uncertain cases sent for human inspection.}
    \label{fig:qc_pipeline}
\end{figure}

\paragraph{Postprocessing for data quality.}
As shown in Fig.~\ref{fig:qc_pipeline}, we use an automated quality-control pipeline that cross-checks tactile recordings against wrist-view observations. For each hand, Gemini 3.7 Flash~\citep{google2026gemini37flash} first identifies grasp intervals, release events, visually confirmed contact fingers, and contact-free periods from the wrist video without access to tactile measurements. This blind visual pass provides an independent reference for evaluating the tactile signals. We reject recordings with clear sensor failures, including whole-hand signal loss during visually confirmed interaction and high-confidence persistent multi-finger signal failures. Missing responses from individual fingers are instead sent for manual review, since finger-level contact can be visually ambiguous. Fingers that respond elsewhere in the episode are not treated as failures when they are silent during a particular grasp. We separately screen contact-free periods for persistent tactile activity and use a second visual check to distinguish sensor noise from unobserved physical contact. More details can be found in Appendix~\ref{app:tactile_qc}.

\section{Experiments}
\label{sec:experiments}

We examine the four questions above through three experimental settings. First, we evaluate cross-dataset zero-shot tactile prediction on EgoTactile and analyze how performance changes with the amount of TouchScale training data. Second, we evaluate whether tactile-supervised pretraining on TouchScale learns transferable visual representations across three downstream action recognition benchmarks under both linear probing and end-to-end fine-tuning. Finally, we evaluate TouchScale as action-free visual--tactile mid-training for $\mathcal{N}_0$-VTLA on four real-world contact-rich manipulation tasks, and further study how robot success and future-tactile prediction change with the amount of TouchScale mid-training data.

\subsection{Tactile Prediction}
\label{sec:tactile_prediction}

We evaluate zero-shot transfer for tactile prediction. We first train the same TouchAnything model~\citep{zhou2026touchanything} on the full EgoTouch training split (16.2 h) and on a size-matched subset of TouchScale ($\sim$16 h), and evaluate both models directly on EgoTactile~\citep{zeng2026egotactile} without any post-training. For EgoTouch, we use the released WiLoR pose estimates~\citep{potamias2025wilor}; for TouchScale, we obtain WiLoR poses from automatically detected hand crops. Ground-truth pressure is normalized to $[0,1]$ by the maximum pressure within each sequence.
Each dataset retains its native tactile layout during training. Because the sensors differ in spatial layout and sensing density, we map their tactile prediction to a common set of 12 anatomical hand regions for evaluation.

We report four region-level metrics: contact intersection-over-union (cIoU), volumetric intersection-over-union (vIoU), center-of-pressure localization error (CoP Loc.), and regional pressure magnitude error (Press. Mag.). These metrics measure regional contact agreement, pressure overlap, pressure localization, and pressure intensity. The anatomical mapping and complete metric definitions are provided in Appendix~\ref{sec:supp_tactile_prediction}.

\begin{table}[t]
\centering
\caption{\textbf{Tactile prediction.}
We compare models trained on EgoTouch and TouchScale under zero-shot transfer to EgoTactile and in-distribution evaluation on each source dataset's unseen split. Zero-shot metrics are averaged over 12 anatomical hand regions.}
\label{tab:tactile_prediction}
\small
\renewcommand{\arraystretch}{1.10}
\setlength{\tabcolsep}{4pt}
\vspace{2mm}

\begin{tabular*}{0.9\columnwidth}{@{\extracolsep{\fill}}llcccc@{}}
\toprule
Evaluation & Training data
& cIoU$\uparrow$
& vIoU$\uparrow$
& CoP Loc.$\downarrow$
& Press. Mag.$\downarrow$ \\
\midrule
\multirow{2}{*}{\shortstack[l]{Zero-shot\\(EgoTactile)}}
& EgoTouch   & 0.134          & 0.185          & 0.444          & 0.184 \\
& TouchScale & \textbf{0.181} & \textbf{0.243} & \textbf{0.416} & \textbf{0.148} \\
\midrule
\multirow{2}{*}{Unseen split}
& EgoTouch   & 0.403          & \textbf{0.302} & 0.272          & 0.095 \\
& TouchScale & \textbf{0.422} & 0.299          & \textbf{0.261} & \textbf{0.061} \\
\bottomrule
\end{tabular*}
\end{table}

Table~\ref{tab:tactile_prediction} shows that, compared to the baseline trained on EgoTouch, training on TouchScale improves zero-shot cIoU from 0.134 to 0.181 and vIoU from 0.185 to 0.243, corresponding to 35.1\% and 31.4\% relative gains, while reducing CoP localization error from 0.444 to 0.416 and regional pressure magnitude error from 0.184 to 0.148. Overall, TouchScale transfers better for contact prediction and spatial pressure estimation across unseen tactile hardware. On the unseen split of each source dataset, TouchScale also achieves higher cIoU (0.422 vs.\ 0.403), lower CoP localization error (0.261 vs.\ 0.272), and lower pressure magnitude error (0.061 vs.\ 0.095), while vIoU remains comparable (0.299 vs.\ 0.302).

\begin{figure}[t]
\centering
\includegraphics[width=\linewidth]{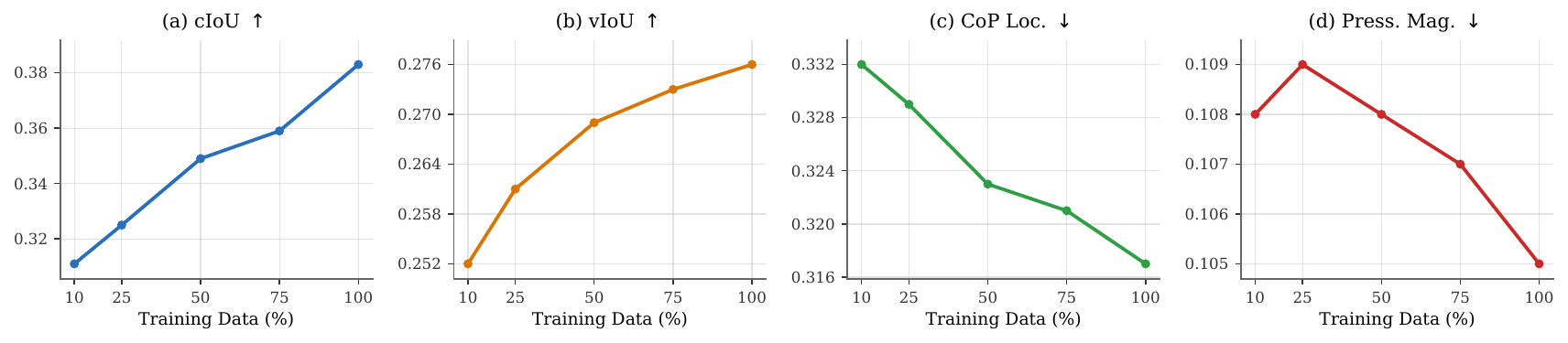}
\caption{\textbf{Scaling of zero-shot tactile prediction.} Increasing TouchScale training data generally improves zero-shot performance on EgoTactile across contact, localization, and pressure metrics.}
\label{fig:tactile_scaling}
\end{figure}

\paragraph{Data scaling.}
We study how TouchScale data scale affects zero-shot tactile prediction while keeping the model and the EgoTactile evaluation protocol fixed. As shown in Fig.~\ref{fig:tactile_scaling}, increasing the training scale from 10\% to 100\% improves cIoU from 0.311 to 0.383 and vIoU from 0.252 to 0.276, while reducing CoP localization error from 0.332 to 0.317 and regional pressure magnitude error from 0.108 to 0.105. Overall, the four metrics show favorable scaling trends, indicating that larger TouchScale training sets improve zero-shot tactile prediction on EgoTactile.

\subsection{Visual Representation Learning}
\label{sec:action_representation}

We next evaluate the transferability of visual representations learned from TouchScale. We pretrain Hiera-B~\citep{ryali2023hiera} separately on TouchScale, OpenTouch~\citep{song2025opentouch}, FEEL~\citep{dessalene2026feel}, and EgoTouch~\citep{zhou2026touchanything}, using the same tactile prediction objective. All encoders are initialized from the same pretrained Hiera-B checkpoint and trained with the same optimization schedule and training budget. We evaluate the resulting encoders on MECCANO~\citep{ragusa2021meccano}, Something-Something V2 (SSv2)~\citep{goyal2017something}, and Ego-Exo4D~\citep{grauman2024egoexo4d} under both linear probing and end-to-end fine-tuning.
Table~\ref{tab:action_representation} shows that TouchScale achieves the highest accuracy across all three downstream benchmarks under both evaluation settings. It reaches an average linear-probe accuracy of 24.45\% and an average fine-tuning accuracy of 49.62\%, outperforming the existing visual--tactile pretraining datasets. These results show that TouchScale learns more transferable visual representations for downstream action recognition.

\begin{table*}[t]
\centering
\caption{\textbf{Downstream action recognition.} TouchScale achieves the highest average accuracy under both linear probing and end-to-end fine-tuning. FT denotes end-to-end fine-tuning.}
\label{tab:action_representation}
\small
\setlength{\tabcolsep}{5pt}
\begin{tabular*}{0.92\linewidth}{@{\extracolsep{\fill}}lcccccccc@{}}
\toprule
& \multicolumn{2}{c}{MECCANO}
& \multicolumn{2}{c}{SSv2}
& \multicolumn{2}{c}{Ego-Exo4D}
& \multicolumn{2}{c}{Avg.} \\
\cmidrule(lr){2-3}
\cmidrule(lr){4-5}
\cmidrule(lr){6-7}
\cmidrule(lr){8-9}
Pretraining Data
& Linear & FT
& Linear & FT
& Linear & FT
& Linear & FT \\
\midrule
OpenTouch
& 28.76 & 42.07
& 20.39 & 62.82
& 19.76 & 39.94
& 22.97 & 48.28 \\

FEEL
& 17.71 & 21.60
& 1.51 & 12.90
& 4.32 & 5.37
& 7.85 & 13.29 \\

EgoTouch
& 20.16 & 24.16
& 1.74 & 29.87
& 5.06 & 11.22
& 8.99 & 21.75 \\
\midrule

\textbf{TouchScale}
& \textbf{28.97} & \textbf{43.71}
& \textbf{22.82} & \textbf{63.27}
& \textbf{21.55} & \textbf{41.89}
& \textbf{24.45} & \textbf{49.62} \\
\bottomrule
\end{tabular*}
\end{table*}

\subsection{Transfer to Robotic Manipulation}
\label{sec:robot_manipulation}

We build on $\mathcal{N}_0$-VTLA~\citep{n0vtla2026}, a vision--tactile--language--action policy for contact-rich manipulation that takes visual observations, language instructions, robot state, and tactile input to predict robot actions. Given the current tactile observation and visual--language context, $\mathcal{N}_0$-VTLA predicts latent tactile tokens that represent the expected tactile change over the coming action chunk and conditions the action expert on these tokens for action prediction. In Stage 1, these latent tactile tokens are trained against a future-tactile target obtained by encoding the tactile change over the same horizon. We follow the Stage-1 future-tactile prediction formulation while adapting the training pipeline to TouchScale before robot post-training, and refer to this stage as TouchScale mid-training. Details are provided in Appendix~\ref{sec:robot_impl}.

\paragraph{TouchScale mid-training and robot post-training.}
Both variants start from the officially released $\mathcal{N}_0$-VTLA checkpoint. The TouchScale variant first undergoes \emph{TouchScale mid-training} and is then post-trained on our robot demonstrations. The baseline skips TouchScale mid-training and is directly post-trained on the same robot demonstrations using the same protocol. The two variants therefore differ only in whether TouchScale mid-training is performed before robot post-training.

\begin{figure*}[t]
\centering
\includegraphics[width=\textwidth,keepaspectratio,clip]{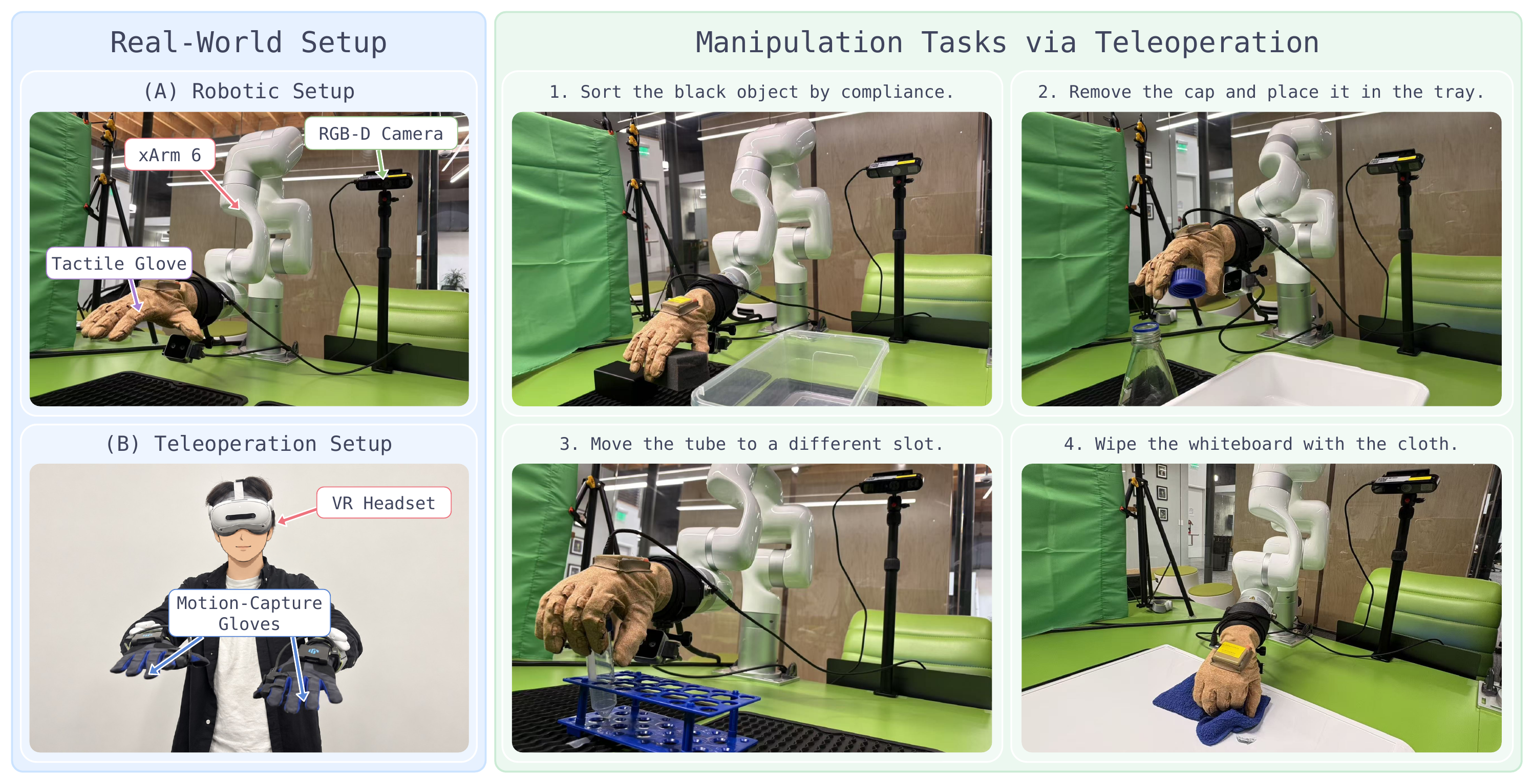}
\caption{\textbf{Real-world robot setup and manipulation tasks.} Left: the xArm6 platform with a BrainCo Revo 2 robotic hand, tactile sensing, and an external RGB-D camera and a wrist camera, together with the teleoperation setup used to collect robot demonstrations. Right: four contact-rich manipulation tasks covering material-dependent sorting, small-object manipulation, precise transfer, and sustained surface contact. We collect 50 teleoperated demonstrations for each task.}
\label{fig:robotic_manipulation}
\end{figure*}

\paragraph{Robot setup and tasks.}
Our real-world setup is shown in Fig.~\ref{fig:robotic_manipulation}. We use an xArm6 robotic arm with a BrainCo Revo 2 robotic hand, a tactile-sensing glove mounted on the hand, an external RGB-D camera, and a wrist camera. We collect 50 demonstrations per task through human teleoperation using a VR headset, a pair of trackers, and motion-capture gloves. We evaluate four contact-rich manipulation tasks with different interaction requirements. In \emph{Soft/Hard Sorting}, the robot identifies the soft object through physical interaction and places it at the target location. In \emph{Bottle-Cap Removal}, the robot picks up a bottle cap and places it into a target container. In \emph{Test-Tube Transfer}, the robot picks up a test tube and transfers it to a target location. In \emph{Whiteboard Wipe}, the robot wipes a marked area on the whiteboard with a cloth while maintaining sustained contact with the surface. For evaluation, we vary the initial configurations of the objects and targets across trials and report the task success rate. Both model variants follow the same evaluation protocol.

\begin{table}[t]
\centering
\caption{\textbf{Real-world robot manipulation.} TouchScale mid-training improves success across all four tasks. Both variants start from the same $\mathcal{N}_0$-VTLA checkpoint and use the same 50 robot demonstrations per task for robot post-training. Each task is evaluated over 20 trials.}
\label{tab:robot_manipulation}
\small
\setlength{\tabcolsep}{2.8pt}
\vspace{2mm}
\begin{tabular*}{0.9\linewidth}{@{\extracolsep{\fill}}cccccc@{}}
\toprule
Mid-training
& \shortstack{Soft/Hard\\Sorting}
& \shortstack{Bottle-Cap\\Removal}
& \shortstack{Test-Tube\\Transfer}
& \shortstack{Whiteboard\\Wipe}
& Avg. \\
\midrule
No
& 10\%
& 40\%
& 30\%
& 10\%
& 22.5\% \\
Yes
& \textbf{60\%}
& \textbf{70\%}
& \textbf{60\%}
& \textbf{40\%}
& \textbf{57.5\%} \\
\bottomrule
\end{tabular*}
\end{table}

\paragraph{Effect of TouchScale mid-training.}
Table~\ref{tab:robot_manipulation} compares $\mathcal{N}_0$-VTLA with and without TouchScale mid-training. TouchScale mid-training improves success on all four real-world contact-rich manipulation tasks, increasing the average success rate from 22.5\% to 57.5\%, an absolute gain of 35 percentage points. The largest improvement is observed on Soft/Hard Sorting, where success increases from 10\% to 60\%. Bottle-Cap Removal improves from 40\% to 70\%, Test-Tube Transfer from 30\% to 60\%, and Whiteboard Wipe from 10\% to 40\%. These consistent gains span material discrimination, small-object manipulation, precise transfer, and continuous-contact manipulation, showing that the benefit of TouchScale mid-training extends across different forms of physical interaction. Overall, these results demonstrate that TouchScale mid-training substantially improves real-world contact-rich robotic manipulation.

\begin{figure*}[t]
\centering
\includegraphics[width=0.8\textwidth]{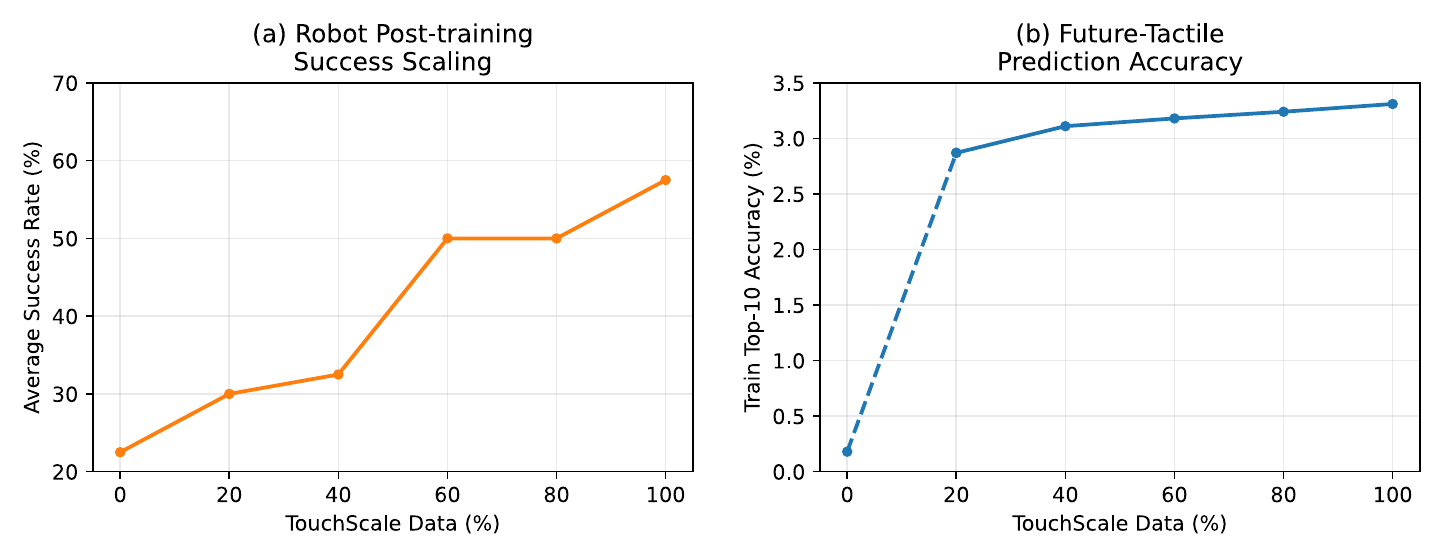}
\caption{\textbf{Scaling of TouchScale mid-training.} Left: average post-training success rate across four real-world manipulation tasks with different amounts of TouchScale mid-training data. Right: future-tactile prediction accuracy on robot data unseen during TouchScale mid-training.}
\label{fig:robot_scaling}
\end{figure*}

\paragraph{Effect of TouchScale data scale on robot policy.}
We next examine how policy performance changes with the amount of TouchScale data used during mid-training. For each data scale, we perform TouchScale mid-training and then post-train the policy on the same robot demonstrations using the same protocol. As shown in Fig.~\ref{fig:robot_scaling} (left), the average success rate across all four real-world tasks increases from 22.5\% without TouchScale mid-training to 30.0\%, 32.5\%, 50.0\%, 50.0\%, and 57.5\% when using 20\%, 40\%, 60\%, 80\%, and 100\% of TouchScale, respectively. The results show an overall positive scaling trend, with larger amounts of human visual--tactile mid-training data leading to stronger downstream robot performance.

\paragraph{Scaling tactile world modeling with TouchScale.}
We explore how TouchScale data scale affects the tactile world modeling during mid-training. Given the current tactile observation and visual--language context, the model predicts latent tactile tokens representing the expected tactile change over the future horizon. We evaluate these predictions on the robot data unseen during TouchScale mid-training, before robot post-training. As shown in Fig.~\ref{fig:robot_scaling} (right), future-tactile prediction accuracy increases from 2.87\% with 20\% of TouchScale to 3.31\% with the full dataset, compared with 0.18\% without TouchScale mid-training. These results show that TouchScale mid-training improves future-tactile prediction on robot observations, while the gain from increasing the amount of human visual--tactile data from 20\% to 100\% is modest. Details are provided in Appendix~\ref{sec:supp_robot_scaling}.

\paragraph{Tactile--action correlation.}
We further examine the relation between future-tactile prediction and action prediction after post-training on robot data. On held-out robot data, we compute the Pearson correlation coefficient between the per-step future-tactile prediction error and action prediction error. With TouchScale mid-training, the two errors show a weak positive correlation ($r=0.119$, $p=0.020$), whereas the baseline shows essentially no correlation ($r=0.003$, $p=0.953$). This result suggests a limited but detectable association between future-tactile prediction and downstream action prediction after mid-training.

\section{Conclusion and Limitations}

We introduce TouchScale, a 500-hour human visual--tactile dataset collected with a consistent sensing and synchronization pipeline. Training on TouchScale improves zero-shot tactile prediction and visual representation learning for downstream action recognition, while mid-training with TouchScale benefits real-world robot manipulation. The scaling results further show overall positive trends as more TouchScale data is used, highlighting the value of large-scale human visual--tactile data for embodied learning. Together, these results validate synchronized human vision and touch as a highly effective supervisory signal for both perception and robotic control. Our current robot evaluation is limited to one platform and four manipulation tasks, and TouchScale does not include action labels. Broader evaluation across robot embodiments and contact-rich tasks remains future work. We also observe only a weak association between future-tactile and action prediction errors, and the role of future-tactile prediction in robot control remains an open question.

\clearpage

\bibliography{paper}

@inproceedings{grauman2022ego4d,
  title     = {Ego4D: Around the World in 3,000 Hours of Egocentric Video},
  author    = {Grauman, Kristen and Westbury, Andrew and Byrne, Eugene and others},
  booktitle = {Proceedings of the IEEE/CVF Conference on Computer Vision and Pattern Recognition},
  pages     = {18995--19012},
  year      = {2022}
}

@inproceedings{grauman2024egoexo4d,
  title     = {Ego-Exo4D: Understanding Skilled Human Activity from First- and Third-Person Perspectives},
  author    = {Grauman, Kristen and Westbury, Andrew and Torresani, Lorenzo and others},
  booktitle = {Proceedings of the IEEE/CVF Conference on Computer Vision and Pattern Recognition},
  pages     = {19383--19400},
  year      = {2024}
}

@inproceedings{hoque2026egodex,
  title     = {EgoDex: Learning Dexterous Manipulation from Large-Scale Egocentric Video},
  author    = {Hoque, Ryan and Huang, Peide and Yoon, David J. and Sivapurapu, Mouli and Zhang, Jian},
  booktitle = {The Fourteenth International Conference on Learning Representations},
  year      = {2026}
}

@inproceedings{kareer2025egomimic,
  title     = {EgoMimic: Scaling Imitation Learning via Egocentric Video},
  author    = {Kareer, Simar and Patel, Dhruv and Punamiya, Ryan and Mathur, Pranay and Cheng, Shuo and Wang, Chen and Hoffman, Judy and Xu, Danfei},
  booktitle = {2025 IEEE International Conference on Robotics and Automation},
  pages     = {13226--13233},
  year      = {2025},
  doi       = {10.1109/ICRA55743.2025.11127989}
}

@article{zheng2026egoscale,
  title   = {{EgoScale}: Scaling Dexterous Manipulation with Diverse Egocentric Human Data},
  author  = {Zheng, Ruijie and Niu, Dantong and Xie, Yuqi and Wang, Jing and Xu, Mengda and Jiang, Yunfan and Casta{\~n}eda, Fernando and Hu, Fengyuan and Tan, You Liang and Fu, Letian and Darrell, Trevor and Huang, Furong and Zhu, Yuke and Xu, Danfei and Fan, Linxi},
  journal = {arXiv preprint arXiv:2602.16710},
  year    = {2026}
}

@inproceedings{liu2025vtdexmanip,
  title     = {VTDexManip: A Dataset and Benchmark for Visual-Tactile Pretraining and Dexterous Manipulation with Reinforcement Learning},
  author    = {Liu, Qingtao and Cui, Yu and Sun, Zhengnan and Li, Gaofeng and Chen, Jiming and Ye, Qi},
  booktitle = {The Thirteenth International Conference on Learning Representations},
  year      = {2025}
}

@article{niu2026trex,
  title   = {{T-Rex}: Tactile-Reactive Dexterous Manipulation},
  author  = {Niu, Dantong and Liu, Zhuoyang and Wang, Zekai and Shao, Boning and Yin, Zhao-Heng and Pai, Anirudh and Sharma, Yuvan and Saravalle, Stefano and Zheng, Ruijie and Wang, Jing and Punamiya, Ryan and Xu, Mengda and Xie, Yuqi and Jiang, Yunfan and Fu, Letian and Kallidromitis, Konstantinos and Gioia, Matteo and Zhang, Junyi and Ge, Jiaxin and Feng, Haiwen and Galasso, Fabio and Zhan, Wei and Chan, David M. and Bai, Yutong and Herzig, Roei and Lei, Jiahui and Li, Fei-Fei and Goldberg, Ken and Malik, Jitendra and Abbeel, Pieter and Zhu, Yuke and Xu, Danfei and Fan, Jim and Darrell, Trevor},
  journal = {arXiv preprint arXiv:2606.17055},
  year    = {2026}
}

@article{song2025opentouch,
  title   = {{OPENTOUCH}: Bringing Full-Hand Touch to Real-World Interaction},
  author  = {Song, Yuxin Ray and Li, Jinzhou and Fu, Rao and Murphy, Devin and Zhou, Kaichen and Shiv, Rishi and Li, Yaqi and Xiong, Haoyu and Owens, Crystal Elaine and Du, Yilun and Luo, Yiyue and Cheng, Xianyi and Torralba, Antonio and Matusik, Wojciech and Liang, Paul Pu},
  journal = {arXiv preprint arXiv:2512.16842},
  year    = {2025}
}

@article{dessalene2026feel,
  title   = {{FEEL} (Force-Enhanced Egocentric Learning): A Dataset for Physical Action Understanding},
  author  = {Dessalene, Eadom and He, Botao and Maynord, Michael and Tussa, Yonatan and Mantripragada, Pavan and Karabati, Yianni and Roy, Nirupam and Aloimonos, Yiannis},
  journal = {arXiv preprint arXiv:2603.15847},
  year    = {2026}
}

@article{zhou2026touchanything,
  title   = {TouchAnything: A Dataset and Framework for Bimanual Tactile Estimation from Egocentric Video},
  author  = {Zhou, Jianyi and Gao, Ziteng and Hong, Feiyang and Liu, Zirui and Zhang, Guannan and Dai, Weisheng and Zhen, Ruichen and Lyu, Chuqiao and Wu, Haotian and Mao, Yinian and Wang, Xushi and Jiang, Yuxiang and Ding, Wenbo and Yang, Shuo},
  journal = {arXiv preprint arXiv:2605.13083},
  year    = {2026}
}

@inproceedings{zeng2026egotactile,
  title     = {EgoTactile: Learning Grasp Pressure for Everyday Objects from Egocentric Video},
  author    = {Zeng, Yuan and Shi, Yujia and Tan, Tiao and Li, Xingting and Qin, Yaqi and Lu, Zongqing and Yang, Wenming and Xue, Jing-Hao and Liao, Qingmin},
  booktitle = {International Conference on Machine Learning},
  year      = {2026}
}

@article{zhang2026human,
  title         = {Human-Centric Transferable Tactile Pre-Training for Dexterous Robotic Manipulation},
  author        = {Zhang, Chi and Cai, Penglin and Xi, Ziheng and Yuan, Haoqi and Luo, Hao and Zhang, Wanpeng and Zheng, Sipeng and Xu, Chaoyi and Lu, Zongqing},
  journal       = {arXiv preprint arXiv:2607.01067},
  year          = {2026},
  eprint        = {2607.01067},
  archivePrefix = {arXiv},
  primaryClass  = {cs.RO}
}

@article{damen2022epickitchens100,
  title   = {Rescaling Egocentric Vision: Collection, Pipeline and Challenges for {EPIC-KITCHENS-100}},
  author  = {Damen, Dima and Doughty, Hazel and Farinella, Giovanni Maria and Furnari, Antonino and Kazakos, Evangelos and Ma, Jian and Moltisanti, Davide and Munro, Jonathan and Perrett, Toby and Price, Will and Wray, Michael},
  journal = {International Journal of Computer Vision},
  volume  = {130},
  number  = {1},
  pages   = {33--55},
  year    = {2022},
  doi     = {10.1007/s11263-021-01531-2}
}

@article{liu2026egotl,
  title   = {{EgoTL}: Egocentric Think-Aloud Chains for Long-Horizon Tasks},
  author  = {Liu, Lulin and Li, Dayou and Liang, Yiqing and Jiang, Sicong and Vijay, Hitesh and Hu, Hezhen and Xu, Xuhai and Liu, Zirui and Shakkottai, Srinivas and Li, Manling and Fan, Zhiwen},
  journal = {arXiv preprint arXiv:2604.09535},
  year    = {2026}
}

@inproceedings{punamiya2026egoverse,
  title     = {{EgoVerse}: An Egocentric Human Dataset for Robot Learning from Around the World},
  author    = {Punamiya, Ryan and Kareer, Simar and Liu, Zeyi and Citron, Josh and Qiu, Ri-Zhao and Cai, Xiongyi and Gavryushin, Alexey and Chen, Jiaqi and Liconti, Davide and Zhu, Lawrence Y. and Aphiwetsa, Patcharapong and Li, Baoyu and Cheluva, Aniketh and Kuppili, Pranav and Liu, Yangcen and Patel, Dhruv and Gao, Aidan and Chung, Hye-Young and Co, Ryan and Zbizika, Renee and others},
  booktitle = {Proceedings of Robotics: Science and Systems},
  year      = {2026}
}

@inproceedings{zhao2025egopressure,
  title     = {{EgoPressure}: A Dataset for Hand Pressure and Pose Estimation in Egocentric Vision},
  author    = {Zhao, Yiming and Kwon, Taein and Streli, Paul and Pollefeys, Marc and Holz, Christian},
  booktitle = {Proceedings of the IEEE/CVF Conference on Computer Vision and Pattern Recognition (CVPR)},
  pages     = {27727--27738},
  year      = {2025},
  doi       = {10.1109/CVPR52734.2025.02582}
}

@article{huang2026htbench,
  title   = {{HT-Bench}: Benchmarking and Learning Dexterous Full-Hand Tactile Representations with Egocentric Vision},
  author  = {Huang, Yuzhe and Wu, Jiaping and Jiang, Jiaming and Lin, Hezhe and Aierken, Aikebaier and Wang, Yunlong and Cheng, Kun and Jiao, Ziyuan and Zhong, Yuanxin},
  journal = {arXiv preprint arXiv:2606.19161},
  year    = {2026}
}

@inproceedings{nair2023r3m,
  title     = {{R3M}: A Universal Visual Representation for Robot Manipulation},
  author    = {Nair, Suraj and Rajeswaran, Aravind and Kumar, Vikash and Finn, Chelsea and Gupta, Abhinav},
  booktitle = {Proceedings of The 6th Conference on Robot Learning},
  pages     = {892--909},
  year      = {2023},
  volume    = {205},
  series    = {Proceedings of Machine Learning Research},
  publisher = {PMLR}
}

@inproceedings{ma2023vip,
  title     = {{VIP}: Towards Universal Visual Reward and Representation via Value-Implicit Pre-Training},
  author    = {Ma, Yecheng Jason and Sodhani, Shagun and Jayaraman, Dinesh and Bastani, Osbert and Kumar, Vikash and Zhang, Amy},
  booktitle = {International Conference on Learning Representations (ICLR)},
  year      = {2023}
}

@inproceedings{qiu2025humanoid,
  title     = {Humanoid Policy $\sim$ Human Policy},
  author    = {Qiu, Ri-Zhao and Yang, Shiqi and Cheng, Xuxin and Chawla, Chaitanya and Li, Jialong and He, Tairan and Yan, Ge and Yoon, David J. and Hoque, Ryan and Paulsen, Lars and Yang, Ge and Zhang, Jian and Yi, Sha and Shi, Guanya and Wang, Xiaolong},
  booktitle = {Proceedings of The 9th Conference on Robot Learning},
  pages     = {2888--2906},
  year      = {2025},
  volume    = {305},
  series    = {Proceedings of Machine Learning Research},
  publisher = {PMLR}
}

@article{yang2025egovla,
  title   = {{EgoVLA}: Learning Vision-Language-Action Models from Egocentric Human Videos},
  author  = {Yang, Ruihan and Yu, Qinxi and Wu, Yecheng and Yan, Rui and Li, Borui and Cheng, An-Chieh and Zou, Xueyan and Fang, Yunhao and Cheng, Xuxin and Qiu, Ri-Zhao and Yin, Hongxu and Liu, Sifei and Han, Song and Lu, Yao and Wang, Xiaolong},
  journal = {arXiv preprint arXiv:2507.12440},
  year    = {2025}
}

@inproceedings{wi2026tactalign,
  title     = {{TactAlign}: Human-to-Robot Policy Transfer via Tactile Alignment},
  author    = {Wi, Youngsun and Yin, Jessica and Xiang, Elvis and Sharma, Akash and Malik, Jitendra and Mukadam, Mustafa and Fazeli, Nima and Hellebrekers, Tess},
  booktitle = {Proceedings of Robotics: Science and Systems},
  year      = {2026},
  address   = {Sydney, Australia},
  month     = jul,
  doi       = {10.15607/RSS.2026.XXII.006}
}

@inproceedings{potamias2025wilor,
  title        = {{WiLoR}: End-to-End 3D Hand Localization and Reconstruction in-the-Wild},
  author       = {Potamias, Rolandos Alexandros and Zhang, Jinglei and Deng, Jiankang and Zafeiriou, Stefanos},
  booktitle    = {2025 IEEE/CVF Conference on Computer Vision and Pattern Recognition (CVPR)},
  pages        = {12242--12254},
  year         = {2025},
  organization = {IEEE}
}

@inproceedings{ragusa2021meccano,
  title     = {The MECCANO Dataset: Understanding Human-Object Interactions from Egocentric Videos in an Industrial-Like Domain},
  author    = {Ragusa, Francesco and Furnari, Antonino and Livatino, Salvatore and Farinella, Giovanni Maria},
  booktitle = {WACV},
  year      = {2021}
}

@inproceedings{goyal2017something,
  title     = {The ``Something Something'' Video Database for Learning and Evaluating Visual Common Sense},
  author    = {Goyal, Raghav and Ebrahimi Kahou, Samira and Michalski, Vincent and Materzynska, Joanna and Westphal, Susanne and Kim, Heuna and Haenel, Valentin and Fruend, Ingo and Yianilos, Peter and Mueller-Freitag, Moritz and others},
  booktitle = {ICCV},
  year      = {2017}
}

@inproceedings{ryali2023hiera,
  title     = {Hiera: A Hierarchical Vision Transformer without the Bells-and-Whistles},
  author    = {Ryali, Chaitanya and Hu, Yuan-Ting and Bolya, Daniel and Wei, Chen and Fan, Haoqi and Huang, Po-Yao and Aggarwal, Vaibhav and Chowdhury, Arkabandhu and Poursaeed, Omid and Hoffman, Judy and Malik, Jitendra and Li, Yanghao and Feichtenhofer, Christoph},
  booktitle = {Proceedings of the 40th International Conference on Machine Learning},
  pages     = {29441--29454},
  year      = {2023},
  volume    = {202},
  series    = {Proceedings of Machine Learning Research},
  publisher = {PMLR}
}

@article{n0vtla2026,
  title   = {{N0-VTLA}: Scaling Vision-Tactile-Language-Action Model with Latent Tactile Tokens},
  author  = {{NeoteAI Team} and {Fudan TEAI Team}},
  journal = {arXiv preprint arXiv:2607.23782},
  year    = {2026}
}

@article{sundaram2019stag,
  title   = {Learning the signatures of the human grasp using a scalable tactile glove},
  author  = {Sundaram, Subramanian and Kellnhofer, Petr and Li, Yunzhu and Zhu, Jun-Yan and Torralba, Antonio and Matusik, Wojciech},
  journal = {Nature},
  volume  = {569},
  number  = {7758},
  pages   = {698--702},
  year    = {2019},
  doi     = {10.1038/s41586-019-1234-z},
  url     = {https://www.nature.com/articles/s41586-019-1234-z}
}

@inproceedings{delpreto2022actionsense,
  title     = {{ActionSense}: A Multimodal Dataset and Recording Framework for Human Activities Using Wearable Sensors in a Kitchen Environment},
  author    = {DelPreto, Joseph and Liu, Chao and Luo, Yiyue and Foshey, Michael and Li, Yunzhu and Torralba, Antonio and Matusik, Wojciech and Rus, Daniela},
  booktitle = {Neural Information Processing Systems (NeurIPS) Track on Datasets and Benchmarks},
  year      = {2022},
  url       = {https://action-sense.csail.mit.edu/}
}

@inproceedings{grady2022pressurevision,
  title     = {{PressureVision}: Estimating Hand Pressure from a Single {RGB} Image},
  author    = {Grady, Patrick and Tang, Chengcheng and Brahmbhatt, Samarth and Twigg, Christopher D. and Wan, Chengde and Hays, James and Kemp, Charles C.},
  booktitle = {European Conference on Computer Vision (ECCV)},
  year      = {2022},
  url       = {https://arxiv.org/abs/2203.10385}
}

@inproceedings{grady2024pressurevisionpp,
  title     = {{PressureVision++}: Estimating Fingertip Pressure From Diverse {RGB} Images},
  author    = {Grady, Patrick and Collins, Jeremy A. and Tang, Chengcheng and Twigg, Christopher D. and Aneja, Kunal and Hays, James and Kemp, Charles C.},
  booktitle = {Proceedings of the IEEE/CVF Winter Conference on Applications of Computer Vision (WACV)},
  pages     = {8698--8708},
  year      = {2024},
  url       = {https://openaccess.thecvf.com/content/WACV2024/html/Grady_PressureVision_Estimating_Fingertip_Pressure_From_Diverse_RGB_Images_WACV_2024_paper.html}
}

@inproceedings{yin2023touchdexterity,
  title     = {Rotating without Seeing: Towards In-hand Dexterity through Touch},
  author    = {Yin, Zhao-Heng and Huang, Binghao and Qin, Yuzhe and Chen, Qifeng and Wang, Xiaolong},
  booktitle = {Proceedings of Robotics: Science and Systems},
  year      = {2023},
  address   = {Daegu, Republic of Korea},
  month     = jul,
  doi       = {10.15607/RSS.2023.XIX.036}
}

@inproceedings{yuan2024robotsynesthesia,
  title     = {Robot Synesthesia: In-Hand Manipulation with Visuotactile Sensing},
  author    = {Yuan, Ying and Che, Haichuan and Qin, Yuzhe and Huang, Binghao and Yin, Zhao-Heng and Lee, Kang-Won and Wu, Yi and Lim, Soo-Chul and Wang, Xiaolong},
  booktitle = {2024 IEEE International Conference on Robotics and Automation (ICRA)},
  pages     = {6558--6565},
  year      = {2024},
  doi       = {10.1109/ICRA57147.2024.10610532}
}

@inproceedings{huang2025vitac,
  title     = {{3D-ViTac}: Learning Fine-Grained Manipulation with Visuo-Tactile Sensing},
  author    = {Huang, Binghao and Wang, Yixuan and Yang, Xinyi and Luo, Yiyue and Li, Yunzhu},
  booktitle = {Proceedings of The 8th Conference on Robot Learning},
  pages     = {2557--2578},
  year      = {2025},
  volume    = {270},
  series    = {Proceedings of Machine Learning Research},
  publisher = {PMLR}
}

@inproceedings{yu2025mimictouch,
  title     = {MimicTouch: Leveraging Multi-modal Human Tactile Demonstrations for Contact-rich Manipulation},
  author    = {Yu, Kelin and Han, Yunhai and Wang, Qixian and Saxena, Vaibhav and Xu, Danfei and Zhao, Ye},
  booktitle = {Proceedings of The 8th Conference on Robot Learning},
  pages     = {4844--4865},
  year      = {2025},
  volume    = {270},
  series    = {Proceedings of Machine Learning Research},
  publisher = {PMLR}
}

@article{zhou2026touchsight,
  title   = {{TouchSight}: Bare-Handed Tactile Prediction from Egocentric Video via Generative Visual Augmentation},
  author  = {Zhou, Danyan and Lu, Jinxuan and Lin, Jiawei and Chen, Tianxing and Lyu, Chuqiao and Ding, Wenbo},
  journal = {arXiv preprint arXiv:2609.20414},
  year    = {2026}
}

@article{zhu2025touchwild,
  title   = {Touch in the Wild: Learning Fine-Grained Manipulation with a Portable Visuo-Tactile Gripper},
  author  = {Zhu, Xinyue and Huang, Binghao and Li, Yunzhu},
  journal = {arXiv preprint arXiv:2507.15062},
  year    = {2025}
}

@article{zhu2026emma,
  title     = {{EMMA}: Scaling Mobile Manipulation via Egocentric Human Data},
  author    = {Zhu, Lawrence Y. and Kuppili, Pranav and Punamiya, Ryan and Aphiwetsa, Patcharapong and Patel, Dhruv and Kareer, Simar and Ha, Sehoon and Xu, Danfei},
  journal   = {IEEE Robotics and Automation Letters},
  volume    = {11},
  number    = {3},
  pages     = {3087--3094},
  year      = {2026}
}

@misc{google2026gemini37flash,
  author       = {{Google DeepMind}},
  title        = {Gemini 3.7 Flash: Model Card},
  year         = {2026},
  month        = aug,
  howpublished = {Google DeepMind model card},
  url          = {https://deepmind.google/models/model-cards/gemini-3-7-flash/},
  note         = {Published August 13, 2026}
}

@article{robotics2025world,
  title={World In Your Hands: A Large-Scale and Open-source Ecosystem for Learning Human-centric Manipulation in the Wild},
  author={{TARS Robotics} and Yuhang Zheng and Jichao Peng and Weize Li and Yupeng Zheng and Xiang Li and Yujie Jin and Julong Wei and Guanhua Zhang and Ruiling Zheng and Ming Cao and Songen Gu and Zhenhong Zou and Kaige Li and Ke Wu and Mingmin Yang and Jiahao Liu and Pengfei Li and Hengjie Si and Feiyu Zhu and Wang Fu and Likun Wang and Ruiwen Yao and Jieru Zhao and Yilun Chen and Wenchao Ding},
  journal={arXiv preprint arXiv:2512.24310},
  year={2025}
}

@article{zhang2026egotac,
  title   = {{EgoTac}: In-the-Wild Tactile Prediction from Egocentric Vision},
  author  = {Zhang, Wenkang and Yuan, Chengbo and Zhang, Zicheng and Cheng, Zhengxue and Gao, Yang},
  journal = {arXiv preprint arXiv:2608.15060},
  year    = {2026}
}
\bibliographystyle{assets/plainnat}

\clearpage

\beginappendix
% \section{Appendix}
% \label{sec:appendix}

\section{Tactile Prediction Evaluation Protocol}
\label{sec:supp_tactile_prediction}

\subsection{Cross-sensor anatomical mapping}
\label{app:cross_sensor_mapping}

TouchScale, EgoTouch, and EgoTactile use tactile sensors with different native layouts and sensing densities. For cross-dataset evaluation, we represent TouchScale on a $32\times44$ hand-shaped grid with 880 valid cells, EgoTouch on its native $21\times21$ hand-shaped grid with 217 valid cells, and EgoTactile on a $17\times19$ hand-shaped grid with 137 valid cells. Direct taxel-wise comparison across these processed layouts is therefore not well defined.

For cross-dataset evaluation, we map all three tactile layouts to the same set of 12 anatomical hand regions. Let
\begin{equation}
\begin{aligned}
\mathcal{F}
&= \{\mathrm{thumb,index,middle,ring,pinky}\},\\
\Omega
&= \left\{f^{\mathrm{tip}},f^{\mathrm{mid/base}}\mid f\in\mathcal{F}\right\}\\
&\quad \cup \{\mathrm{palm},\mathrm{thumb\mbox{-}base}\},
\qquad |\Omega|=12.
\end{aligned}
\label{eq:shared_anatomical_set}
\end{equation}
The shared space contains five fingertip regions, five finger middle/base regions, the palm, and the thumb base.

For dataset $d$ with $N_d$ valid native taxels, we define a fixed dataset-specific lookup
\begin{equation}
q_d:\{1,\ldots,N_d\}\rightarrow\Omega\cup\{\varnothing\},
\end{equation}
where $q_d(m)$ gives the anatomical region associated with native taxel $m$, and $\varnothing$ denotes taxels excluded from evaluation. The corresponding binary assignment matrix is
\begin{equation}
\mathbf{M}^{(d)} \in \{0,1\}^{12\times N_d},
\qquad
M^{(d)}_{r,m}
=
\mathbb{I}\!\left[q_d(m)=r\right].
\label{eq:supp_region_mapping}
\end{equation}

\begin{figure*}[!tbp]
    \centering
    \includegraphics[width=\textwidth]{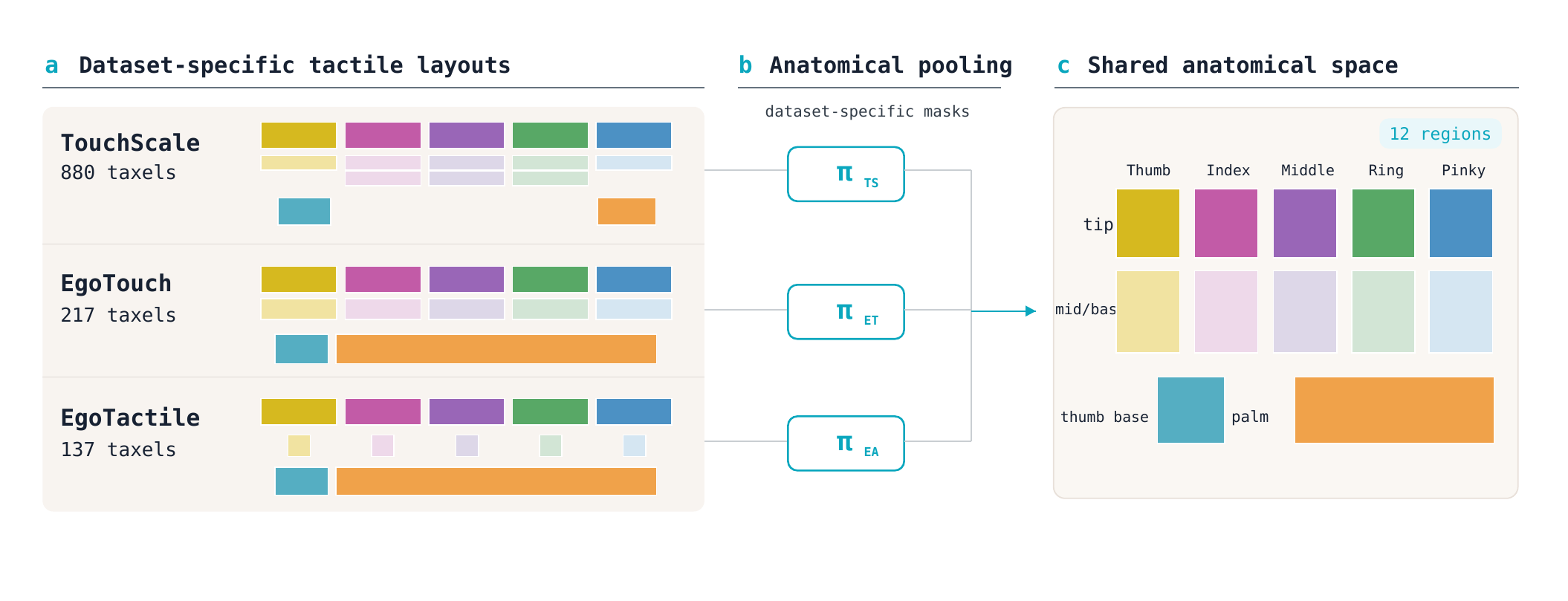}
    \caption{\textbf{Cross-sensor anatomical mapping.} Dataset-specific anatomical assignments pool the TouchScale, EgoTouch, and EgoTactile tactile layouts into the same 12 anatomical regions for cross-dataset evaluation.}
    \label{fig:cross_sensor_mapping}
\end{figure*}

The lookup is constructed from the anatomical location of each sensing taxel in the native tactile glove layout. Finger middle and base taxels are merged into the corresponding $\mathrm{mid/base}$ region when they are represented separately in the native sensor definition. Fingertip, palm, and thumb-base taxels retain their anatomical identities. Left- and right-hand layouts are represented in the same hand-centric orientation before applying the mapping. Figure~\ref{fig:cross_sensor_mapping} illustrates the dataset-specific assignments and the resulting shared anatomical space.

Importantly, this mapping does not resize or interpolate the native tactile grids. Each dataset retains its native tactile representation during training, and anatomical pooling is applied only for cross-dataset evaluation. The same mapping is applied to predictions and ground truth.

Let $p_{t,m}$ and $\hat{p}_{t,m}$ denote the normalized ground-truth and predicted pressure at native taxel $m$ and time $t$, respectively. Ground-truth pressure is normalized to $[0,1]$ by the maximum pressure within each sequence, while the model prediction is already bounded to $[0,1]$. For clarity, we omit the dataset superscript on $\mathbf{M}$ in the following definitions.

\subsection{Region-level tactile representation}

Using the anatomical assignment above, we aggregate each native tactile map into region-level quantities that are independent of the number of sensing taxels within each region.

The binary contact state of region $r$ is defined as
\begin{equation}
c_{t,r}(p)
=
\mathbb{I}
\left[
\frac{
\sum_m M_{r,m}\mathbb{I}[p_{t,m}>\tau]
}{
\sum_m M_{r,m}
}
>
\rho
\right],
\qquad
\tau = 0.1,\quad \rho = 0.2,
\label{eq:supp_region_contact}
\end{equation}
where $\tau=0.1$ is the taxel-level contact threshold applied to normalized pressure, and $\rho=0.2$ is the minimum fraction of active taxels required for a region to be considered in contact. For ground truth, $\tau=0.1$ corresponds to 10\% of the per-sequence maximum pressure. Using the fraction of active taxels rather than their absolute number reduces the effect of different sensing densities across datasets.

The mean pressure within region $r$ is
\begin{equation}
\mu_{t,r}(p)
=
\frac{
\sum_m M_{r,m}p_{t,m}
}{
\sum_m M_{r,m}
}.
\label{eq:supp_region_pressure}
\end{equation}

We compute the regional center of pressure (CoP) as
\begin{equation}
\mathbf{g}_{t,r}(p)
=
\frac{
\sum_m M_{r,m}p_{t,m}\mathbf{x}_m
}{
\sum_m M_{r,m}p_{t,m}+\epsilon
},
\label{eq:supp_region_cop}
\end{equation}
where $\mathbf{x}_m\in[0,1]^2$ denotes the normalized spatial coordinate of native taxel $m$ in a shared hand-centric coordinate frame, and $\epsilon$ is a small constant for numerical stability.

The regional pressure magnitude is defined as
\begin{equation}
s_{t,r}(p)
=
\frac{
\sum_m M_{r,m}p_{t,m}^{2}
}{
\sum_m M_{r,m}p_{t,m}+\epsilon
}.
\label{eq:supp_region_magnitude}
\end{equation}
The total regional pressure, used only to select which region--time pairs are scored for the center-of-pressure metrics, is
\begin{equation}
S_{t,r}(p)=\sum_{m} M_{r,m}\,p_{t,m}.
\end{equation}
We call a region--time pair $(t,r)$ evaluable, written as $(t,r)\in\mathcal{E}$, when both the prediction grid and the ground-truth grid contribute at least one valid taxel to region $r$ at time $t$. For the evaluation on the single-hand EgoTactile dataset, every pair is evaluable. This restriction only takes effect when the dataset-specific layout has no taxel in a region. The same region-level quantities are computed from the predicted pressure $\hat{p}$.

\subsection{Evaluation metrics}

We evaluate tactile prediction using four region-level metrics. Each metric is computed per sequence over that sequence's frames and then macro-averaged over the $N$ test sequences. We use $\langle\cdot\rangle_{\mathrm{seq}}$ to denote this per-sequence macro-average. Within a sequence, the average over regions is taken only over regions with a non-empty denominator, defined separately for each metric below.

\paragraph{Contact IoU.}
We compute contact intersection-over-union over time for each anatomical region and average the result across regions:
\begin{equation}
\mathrm{cIoU}=\Bigg\langle
\frac{1}{|\mathcal{R}^{\mathrm{c}}|}\sum_{r\in\mathcal{R}^{\mathrm{c}}}
\frac{\sum_{t:(t,r)\in\mathcal{E}} c_{t,r}(\hat{p})\,c_{t,r}(p)}
     {\sum_{t:(t,r)\in\mathcal{E}} \mathbb{I}\!\left[c_{t,r}(\hat{p})+c_{t,r}(p)>0\right]+\epsilon}
\Bigg\rangle_{\mathrm{seq}},
\end{equation}
where the average runs over the regions contacted in at least one evaluable frame,
\begin{equation}
\mathcal{R}^{\mathrm{c}}=\Big\{\,r:\textstyle\sum_{t:(t,r)\in\mathcal{E}}\mathbb{I}\!\left[c_{t,r}(\hat{p})+c_{t,r}(p)>0\right]>0\,\Big\}.
\end{equation}

\paragraph{Volumetric IoU.}
Volumetric IoU (vIoU) measures the overlap between predicted and ground-truth regional pressure magnitudes over time:
\begin{equation}
\mathrm{vIoU}=\Bigg\langle
\frac{1}{|\mathcal{R}^{\mathrm{v}}|}\sum_{r\in\mathcal{R}^{\mathrm{v}}}
\frac{\sum_{t:(t,r)\in\mathcal{E}} \min\!\big(\mu_{t,r}(\hat{p}),\mu_{t,r}(p)\big)}
     {\sum_{t:(t,r)\in\mathcal{E}} \max\!\big(\mu_{t,r}(\hat{p}),\mu_{t,r}(p)\big)+\epsilon}
\Bigg\rangle_{\mathrm{seq}},
\end{equation}
where the average runs over the regions with non-zero pooled pressure,
\begin{equation}
\mathcal{R}^{\mathrm{v}}=\Big\{\,r:\textstyle\sum_{t:(t,r)\in\mathcal{E}}\max\!\big(\mu_{t,r}(\hat{p}),\mu_{t,r}(p)\big)>0\,\Big\}.
\end{equation}

\paragraph{CoP localization error.}
The center of pressure is meaningful only for regions where the predicted or ground-truth regional pressure is non-trivial. For each region $r$, let
\begin{equation}
\mathcal{A}_r=\big\{\,t:(t,r)\in\mathcal{E}\ \text{and}\ \big(S_{t,r}(\hat{p})>\tau\ \text{ or }\ S_{t,r}(p)>\tau\big)\big\}
\end{equation}
denote the frames in which the total regional pressure $S_{t,r}$ of the prediction or ground truth exceeds the threshold $\tau$, and let
\begin{equation}
\mathcal{R}^{\mathcal{A}}=\big\{\,r:\mathcal{A}_r\neq\varnothing\,\big\}
\end{equation}
denote the regions evaluated in at least one such frame.
The CoP localization error is
\begin{equation}
E_{\mathrm{CoP}}=\Bigg\langle
\frac{1}{|\mathcal{R}^{\mathcal{A}}|}\sum_{r\in\mathcal{R}^{\mathcal{A}}}
\frac{1}{|\mathcal{A}_r|}\sum_{t\in\mathcal{A}_r}\big\lVert \hat{g}_{t,r}-g_{t,r}\big\rVert_2
\Bigg\rangle_{\mathrm{seq}}.
\end{equation}

\paragraph{Regional pressure magnitude error.}
Over the same region--time sets $\mathcal{A}_r$ and regions $\mathcal{R}^{\mathcal{A}}$, the regional pressure magnitude error measures the absolute difference between the predicted and ground-truth regional pressure magnitudes:
\begin{equation}
E_{\mathrm{mag}}=\Bigg\langle
\frac{1}{|\mathcal{R}^{\mathcal{A}}|}\sum_{r\in\mathcal{R}^{\mathcal{A}}}
\frac{1}{|\mathcal{A}_r|}\sum_{t\in\mathcal{A}_r}\big| s_{t,r}(\hat{p})-s_{t,r}(p)\big|
\Bigg\rangle_{\mathrm{seq}}.
\end{equation}

These four metrics correspond to cIoU, vIoU, CoP Loc., and Press. Mag. in the main paper. Higher cIoU and vIoU indicate better performance, while lower CoP Loc. and Press. Mag. indicate better performance.

\subsection{Qualitative Results of Tactile Prediction}

\begin{figure}[t]
    \centering
    \includegraphics[width=\linewidth]{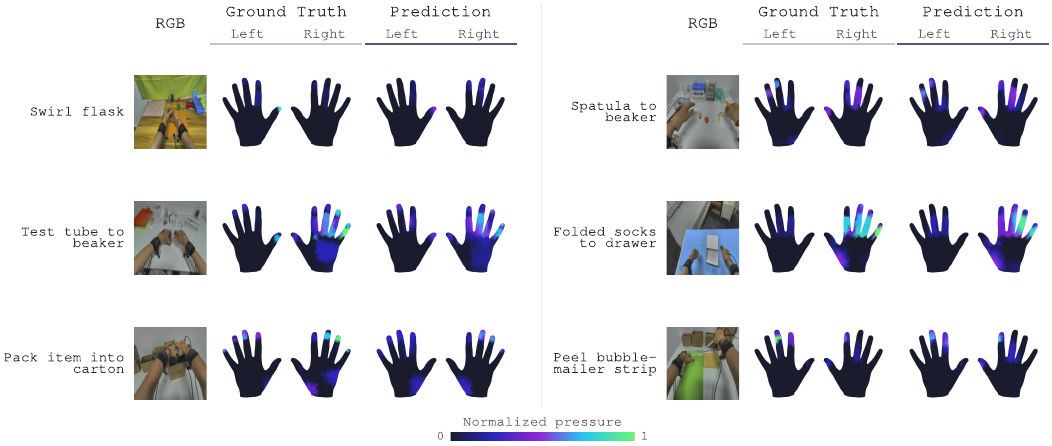}
    \caption{\textbf{Qualitative tactile prediction.} Across diverse human interactions, the predictions closely reproduce the ground-truth contact regions and pressure distributions for both hands, showing consistent tactile prediction across objects and interaction patterns.}
    \label{fig:tactile_prediction}
\end{figure}

Figure~\ref{fig:tactile_prediction} shows representative tactile predictions from the model trained on TouchScale. Across diverse contact-rich interactions, the predictions recover the dominant contact regions on both hands and closely follow the spatial distribution of the ground-truth pressure. The model captures both sparse fingertip contacts and broader multi-region contact patterns, showing consistent tactile prediction across different objects and interaction patterns.

\FloatBarrier

\section{Tactile Glove Details}
\label{app:tactile_glove}

\begin{figure}[!tbp]
    \centering
    \includegraphics[width=0.8\linewidth]{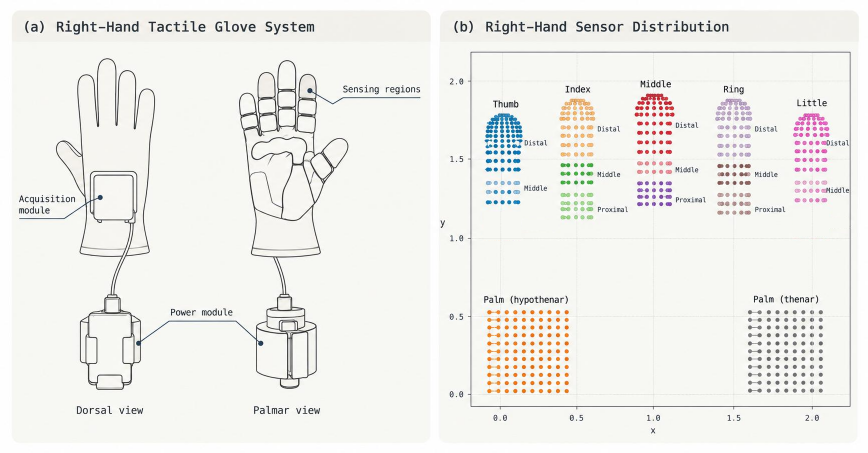}
    \caption{\textbf{Tactile glove hardware.}
The Tachin Glove used in TouchScale contains an acquisition module on the dorsal side, while tactile sensing regions cover the five fingers and palm on the palmar side. The glove is connected to an external power module for wearable tactile recording.}
    \label{fig:tactile_glove}
\end{figure}

We use flexible tactile sensing gloves (Tachin Glove) for full-hand tactile data collection. Each glove contains 880 sensing taxels distributed over the five fingers and palm, as illustrated in Fig.~\ref{fig:tactile_glove}. Each sensing unit measures $1.5\times1.5$\,mm, and the glove provides a spatial resolution below 2\,mm. The sensing layer records normal tactile pressure. The measurement range is 0.2--30 N/cm$^2$, with a resolution of 0.05 N/cm$^2$.

The sensing layer is connected to an acquisition module mounted on the back of the hand and an external power module. The glove supports wireless acquisition at 50\,Hz over WiFi and wired acquisition at 100\,Hz through USB Type-C. The battery module supports more than 10 hours of continuous operation. Together, these components form the wearable tactile acquisition system used throughout data collection.

\FloatBarrier

\section{Tactile Data Quality Control}
\label{app:tactile_qc}

Overfit Lab is our end-to-end data solutions provider; Noitom Robotics handles the data collection layer of the stack. Our data providers obtain informed consent from participants prior to data collection and perform an initial vendor-side quality check before delivery to us. We then apply our own quality-control procedure at the batch, episode, and hand levels. At the batch level, we first check that the required camera streams, timestamps, tactile recordings, and task metadata are present for each episode. We also compute basic integrity statistics, including agreement between video frame counts and timestamps, irregular frame intervals, consistency of recording duration across streams, and interruptions in tactile sampling. These statistics are stored for inspection but are not used directly to determine data quality. We additionally render a synchronized visualization of the camera streams, tactile hand maps, and pressure traces for later review. Each valid episode is then processed independently for the left and right hands. The final episode label is determined by the more severe hand-level result: \textsc{Reject}, \textsc{Review}, or \textsc{Pass}.

\paragraph{Missing tactile signals.}
For each hand, we first test whether all tactile measurements remain below a small response threshold throughout the episode. We use a normalized threshold of $0.03$. A fully silent hand is not considered faulty by itself, since the hand may simply remain unused. To determine whether contact is expected, we run Gemini-3.7-Flash on the corresponding wrist-view video at 2\,fps without providing any tactile information. The model identifies grasp and release intervals, fingers that are visibly involved in each contact, fingers whose contact state is uncertain, and intervals in which the hand is visually free. This blind visual pass provides an independent reference for evaluating the tactile recordings. A fully silent hand is rejected only when the visual analysis confirms that the hand interacts with an object.

For finger-level analysis, tactile measurements are grouped by finger, using the maximum response over the tip, middle, and base regions, with an additional trace for the palm. Values below the response threshold are treated as zero. We evaluate visually confirmed contacts on the thumb, index, middle, and ring fingers; the little finger and palm are excluded from the missing-signal decision because their contact state is less reliable from the wrist view. For each contact interval, we identify confirmed contact fingers whose tactile response remains zero. If all confirmed contact fingers in an interval are silent, at least two such fingers are involved, and these fingers also remain unresponsive throughout the episode, we treat the case as a high-confidence sensor failure and assign \textsc{Reject}. A finger that remains unresponsive throughout the episode without this multi-finger evidence results in \textsc{Review}. If the same finger responds elsewhere in the episode, the missing response is treated as uncertain rather than as a sensor failure.

\paragraph{Tactile noise.}
We separately examine tactile activity during intervals that the visual pass identifies as contact-free. Continuous responses above the same threshold lasting at least 0.3\,s are collected as noise candidates. Candidates are ranked by their duration and peak response, and the five strongest are retained for verification. For each candidate, Gemini-3.7-Flash receives a short clip containing the enlarged tactile visualization alongside the synchronized wrist view, with 1.5\,s of context before and after the candidate at 6\,fps. Unlike the first VLM call, this second call has access to both visual and tactile information and is used only to determine whether the apparent tactile activity occurs while the hand is truly free of contact. Candidates explained by visible contact are discarded; confirmed contact-free responses are recorded as tactile noise.

Tactile noise does not independently determine the \textsc{Reject}/\textsc{Review}/\textsc{Pass} label. We record its timing, cumulative duration, fraction of visually free time, and affected fingers for quality analysis. For each episode, the two hands are evaluated independently, and the more severe hand-level result determines the final episode label. Rejected and review cases are exported with corresponding visualizations for manual inspection.

\FloatBarrier

\section{Future-Tactile Prediction Accuracy}
\label{sec:supp_robot_scaling}

We evaluate whether the tactile predictor learned during Stage 1 transfers from human visual--tactile data to robot observations. For each sample $i$, the Stage-1 predictor produces latent tactile tokens $\hat{Z}_i$ representing the expected tactile change over the future horizon, while the observed tactile change over the same horizon is encoded to obtain the target representation $Z_i^*$.

We mean-pool and $\ell_2$-normalize both representations,
\begin{equation}
\hat{\mathbf{z}}_i =
\frac{\operatorname{MeanPool}(\hat{\mathbf{Z}}_i)}
{\|\operatorname{MeanPool}(\hat{\mathbf{Z}}_i)\|_2},
\qquad
\mathbf{z}^{*}_i =
\frac{\operatorname{MeanPool}(\mathbf{Z}^{*}_i)}
{\|\operatorname{MeanPool}(\mathbf{Z}^{*}_i)\|_2},
\end{equation}
and compute their cosine similarity as
\begin{equation}
s_{ij} = \hat{\mathbf{z}}_i^\top \mathbf{z}^{*}_j.
\end{equation}
An accurate prediction should assign higher similarity to its corresponding target representation than to the tactile-change targets from other samples.

\paragraph{Metrics.}
For each predicted representation $\hat{\mathbf{z}}_i$, we rank all target representations according to cosine similarity. Top-$K$ accuracy measures the fraction of samples for which the corresponding future tactile target appears among the $K$ highest-ranked candidates:
\begin{equation}
\mathrm{Top}\text{-}K =
\frac{1}{N}
\sum_{i=1}^{N}
\mathbb{I}
\left[
\operatorname{rank}
\left(
\mathbf{z}^{*}_i \mid \hat{\mathbf{z}}_i
\right)
\leq K
\right].
\end{equation}
Higher values indicate more accurate future-tactile prediction. We use the Top-10 accuracy as the primary metric in the main paper. We additionally report mean reciprocal rank (MRR),
\begin{equation}
\mathrm{MRR} =
\frac{1}{N}
\sum_{i=1}^{N}
\frac{1}{
\operatorname{rank}
\left(
\mathbf{z}^{*}_i \mid \hat{\mathbf{z}}_i
\right)
},
\end{equation}
where higher values indicate that the corresponding future tactile target is ranked closer to the top. Finally, we report the contrastive loss over the full candidate set,
\begin{equation}
\mathcal{L}_{\mathrm{pool}} =
-\frac{1}{N}
\sum_{i=1}^{N}
\log
\frac{
\exp(s_{ii}/\tau)
}{
\sum_{j=1}^{N}\exp(s_{ij}/\tau)
},
\end{equation}
where $\tau$ is the temperature. Lower Pool NCE indicates better separation between the corresponding future tactile representation and the remaining candidates.

\paragraph{Evaluation protocol.}
We evaluate all Stage-1 checkpoints on the same downstream robot trajectories. The Train split contains 55 episodes and 5,469 valid samples after temporal subsampling, while the Val split contains 13 episodes and 3,070 valid samples. Their union forms a pooled set of 8,539 samples. None of these trajectories is used during Stage-1 mid-training. We use Train Top-10 as the primary metric because it evaluates future-tactile prediction on the robot-data distribution used for policy post-training, while the Val and pooled results are reported as additional reference.

\begin{table*}[t]
\centering
\caption{\textbf{Future-tactile prediction accuracy on robot observations.} TouchScale mid-training improves prediction accuracy, with a modest positive scaling trend in Train Top-10.}
\label{tab:supp_robot_accuracy}

\vspace{4pt}

\begin{tabular}{@{}cccccccc@{}}
\toprule
& \multicolumn{2}{c}{Train}
& \multicolumn{2}{c}{Val}
& \multicolumn{3}{c}{Pooled} \\
\cmidrule(lr){2-3}
\cmidrule(lr){4-5}
\cmidrule(lr){6-8}
Mid-training Data
& Top-10$\uparrow$
& MRR$\uparrow$
& Top-10$\uparrow$
& MRR$\uparrow$
& Top-10$\uparrow$
& MRR$\uparrow$
& Pool NCE$\downarrow$ \\
\midrule
None
& 0.18
& 0.0016
& 0.62
& 0.0039
& 0.14
& 0.0012
& 9.031 \\

20\%
& 2.87
& 0.0083
& 2.93
& 0.0165
& 2.34
& 0.0063
& 8.890 \\

40\%
& 3.11
& 0.0090
& 2.80
& 0.0170
& 2.63
& 0.0070
& 8.889 \\

60\%
& 3.18
& \textbf{0.0108}
& 3.13
& 0.0154
& 2.58
& \textbf{0.0075}
& 8.884 \\

80\%
& 3.24
& 0.0094
& 3.13
& 0.0158
& \textbf{2.69}
& 0.0066
& 8.890 \\

100\%
& \textbf{3.31}
& 0.0094
& \textbf{4.17}
& \textbf{0.0176}
& 2.49
& 0.0064
& \textbf{8.883} \\
\bottomrule
\label{tab:future_tactile_prediction}
\end{tabular}
\end{table*}

Without TouchScale Stage-1 training, Train Top-10 accuracy is 0.18\%, approximately the random-chance level. TouchScale Stage-1 training raises the accuracy to 2.87--3.31\%. As the amount of TouchScale data increases from 20\% to 100\%, Train Top-10 improves from 2.87\% to 3.31\%. The Val and pooled results are reported in Table~\ref{tab:future_tactile_prediction} as additional reference. Overall, TouchScale Stage-1 training improves Train Top-10 over the checkpoint without human mid-training, while increasing the TouchScale scale from 20\% to 100\% provides a modest additional gain.

\FloatBarrier

\section{Real-world Contact-rich Robotic Manipulation}
\label{sec:supp_robot_qualitative}

\subsection{Qualitative Results}

\begin{figure}[!tbp]
    \centering
    \includegraphics[width=\linewidth]{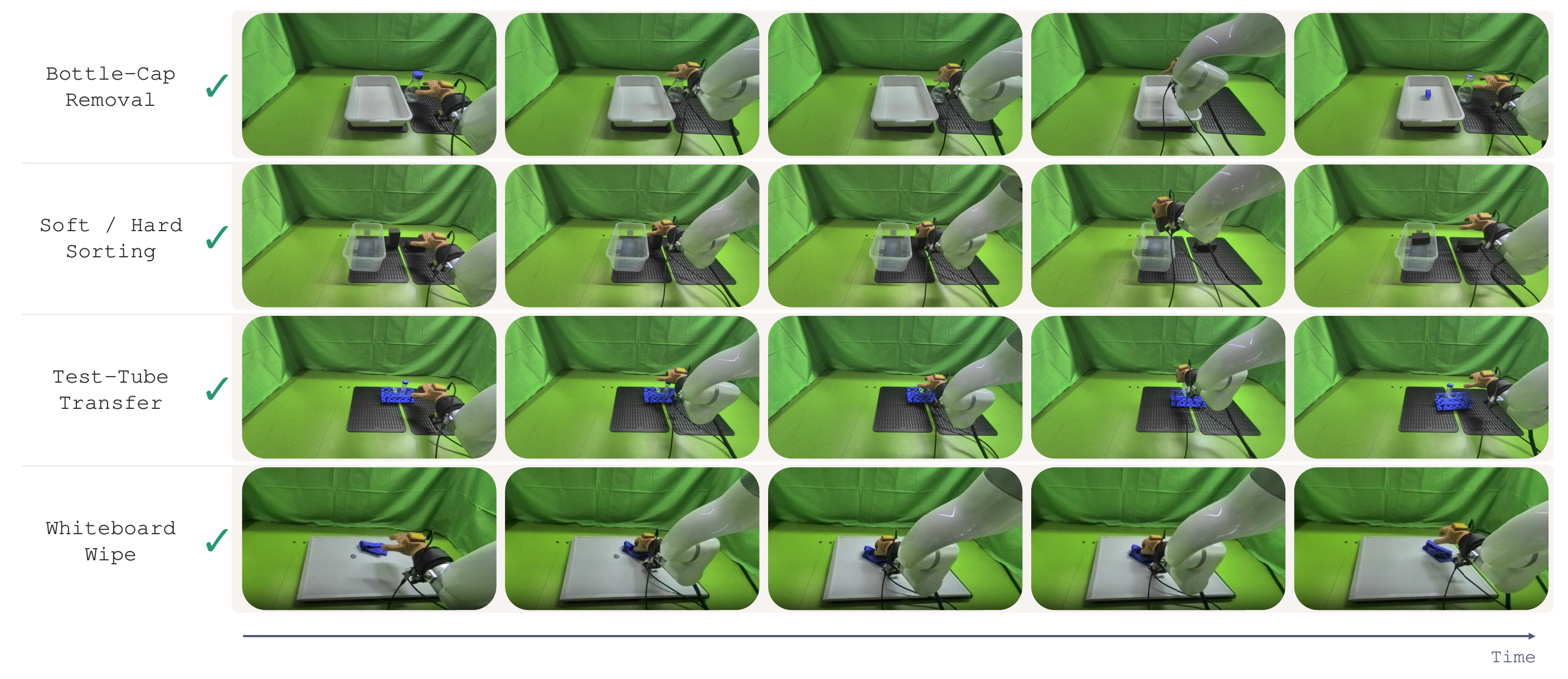}
    \caption{\textbf{Successful executions with TouchScale mid-training.} Representative real-world rollouts across the four manipulation tasks. Each row shows the progression from initial interaction to successful task completion.}
    \label{fig:real_world_success}
\end{figure}

We provide qualitative examples to show the robotic execution. Figure~\ref{fig:real_world_success} shows representative successful executions after TouchScale mid-training across all four tasks: Bottle-Cap Removal, Soft/Hard Sorting, Test-Tube Transfer, and Whiteboard Wipe. Each row follows one rollout over time and illustrates the progression from establishing physical interaction to completing the manipulation goal. The examples cover different contact requirements, including object compliance, small-object handling, precise transfer, and continuous contact during wiping.

\begin{figure}[t]
    \centering
    \includegraphics[width=\linewidth]{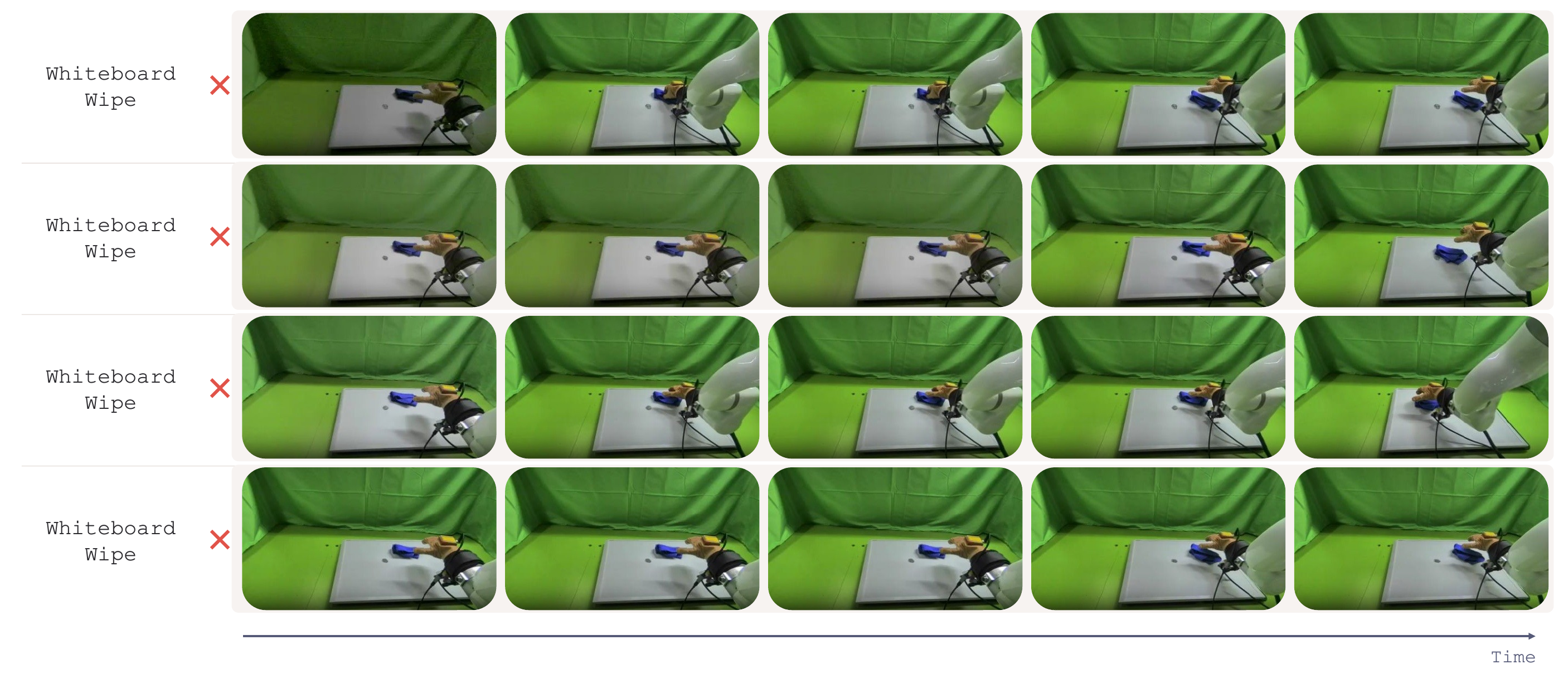}
    \caption{\textbf{Failure cases without TouchScale mid-training.} Representative Whiteboard Wipe rollouts from the baseline $\mathcal{N}_0$-VTLA policy that fail to complete the manipulation task.}
    \label{fig:real_world_failure}
\end{figure}

Figure~\ref{fig:real_world_failure} shows representative Whiteboard Wipe failures from the baseline $\mathcal{N}_0$-VTLA policy without TouchScale mid-training. Although the baseline can initiate interaction with the whiteboard, these rollouts do not successfully complete the wiping task. Together with the quantitative results in the main paper, these examples illustrate the improvement in contact-rich manipulation after TouchScale mid-training.

\FloatBarrier

\subsection{Implementation Details}
\label{sec:robot_impl}

\paragraph{TouchScale mid-training.}
We initialize from the officially released $\mathcal{N}_0$-VTLA checkpoint and follow its Stage-1 future-tactile prediction formulation while adapting the training pipeline to TouchScale. During TouchScale mid-training, the vision--language--action backbone is frozen and no human action labels are used. We update only the tactile projection layer, the tactile predictor, and an auxiliary tactile reconstruction head. The predictor uses the \texttt{tactile\_kv} architecture with five latent tactile tokens, matching the released checkpoint configuration. Given the current tactile observation together with visual--language context, the predictor produces latent tactile tokens that represent the expected tactile change over a future horizon of $H=50$ frames. 
% We use the normal-force channel of TouchScale for this stage.

For each training sample, the future-tactile target is constructed from the tactile change between the current observation and the observation at $t+H$. This tactile change is encoded with the tactile encoder, and targets from the available tactile views are averaged to obtain the target latent representation $z^{*}$. The target branch is stop-gradient during Stage-1 training. We mean-pool and $\ell_2$-normalize the predicted and target latent tactile representations and optimize a symmetric InfoNCE objective over the global training batch. In addition, a two-layer reconstruction head maps the predicted latent tactile tokens to an $8\times8$ representation of the corresponding future tactile change. The Stage-1 objective is
\begin{equation}
\mathcal{L}_{\mathrm{mid}}
=
\mathcal{L}_{\mathrm{NCE}}
+
\lambda_{\mathrm{rec}}\mathcal{L}_{\mathrm{rec}},
\end{equation}
where $\mathcal{L}_{\mathrm{rec}}$ is an $\ell_1$ reconstruction loss and $\lambda_{\mathrm{rec}}=0.5$. We use a temperature of $0.07$ for the contrastive objective.

\paragraph{Training configuration.}
We optimize the trainable tactile pathway with AdamW using a global batch size of 64 and gradient clipping at 1.0. The learning rate is warmed up for 500 steps to $1\times10^{-4}$ and then decayed to $1\times10^{-5}$ by 20K steps. Approximately 123.8M parameters are updated during TouchScale mid-training, while the remaining 3.71B parameters of the base policy remain frozen. The vision--language backbone is kept in evaluation mode and evaluated without gradient computation throughout this stage.

\paragraph{Multimodal alignment.}
TouchScale recordings retain their original timestamps in the released dataset. For $\mathcal{N}_0$-VTLA mid-training, the three RGB streams and two tactile streams are converted to a common 30\,Hz training timeline using timestamp-based nearest-neighbor association. This conversion is performed only when constructing the model training samples and does not modify the original TouchScale recordings. We use fixed tactile normalization statistics estimated from the training split and apply the same normalization throughout mid-training.

\paragraph{Robot post-training.}
After TouchScale mid-training, the resulting checkpoint is post-trained on our robotic demonstrations using the $\mathcal{N}_0$-VTLA robot adaptation pipeline. The baseline starts from the same released checkpoint but skips TouchScale mid-training. Both variants are post-trained with the same 50 robot demonstrations per task and the same optimization and data-processing protocol. Thus, the only difference between the two variants is whether the tactile predictor receives action-free human visual--tactile mid-training on TouchScale before robot post-training.

\section{3D Object Reconstruction}
\label{app:object_reconstruction}

For objects in TouchScale, we record short object-centric RGB videos by moving the camera around each object to capture it from multiple viewpoints. We select several frames in which the object is fully visible and segment the foreground object from the background. The resulting multi-view RGB images are provided to the Meshy Multi-Image-to-3D API to reconstruct a 3D model for each object. These reconstructed object models are provided together with TouchScale.

\end{document}